\documentclass[letterpaper, 10 pt, conference]{ieeeconf}  
\IEEEoverridecommandlockouts

\usepackage[utf8]{inputenc} 
\usepackage[T1]{fontenc}
\usepackage{amsmath}
\usepackage{amsfonts}
\usepackage{float}
\usepackage{graphicx}
\usepackage{booktabs}
\usepackage[font=small]{caption}
\usepackage{subcaption}
\usepackage[ruled]{algorithm2e}
\usepackage{booktabs} 
\usepackage{enumerate}
\usepackage{times}
\usepackage{soul}
\usepackage{url}
\usepackage{hyperref}
\usepackage{bm}
\usepackage{adjustbox}
\usepackage{xcolor}
\usepackage{placeins}
\usepackage[utf8]{inputenc}
\usepackage{booktabs}
\usepackage{array, multirow}
\usepackage{cleveref}
\usepackage{tikz}
\usepackage{wrapfig}
\usepackage{adjustbox}
\usepackage{multirow}
\usepackage{amssymb}
\usepackage{pifont}

\newcommand{\bb}{\mathbf}  
\begin{document}

\newcommand\mycommfont[1]{\footnotesize\rmfamily\textcolor{blue}{#1}}
\usetikzlibrary{arrows.meta}
\usetikzlibrary{positioning}
\tikzstyle{decision} = [diamond, draw, fill=blue!20, 
    text width=6em, text badly centered, node distance=3cm, inner sep=0pt]
\tikzstyle{block} = [rectangle, draw, fill=gray!10, 
    text width=10em, very thick, text centered, rounded corners, minimum height=2.2em]
\tikzstyle{line} = [draw, -{latex[scale=15.0]}]
\tikzstyle{cloud} = [draw, ellipse,fill=red!20, node distance=3cm,
    minimum height=2em]
\setlength{\fboxrule}{1pt}
\setlength{\fboxsep}{0pt}

\newcommand{\cmark}{\ding{51}}%
\newcommand{\xmark}{\ding{55}}%

\newtheorem{innercustomthm}{Theorem}
\newenvironment{customthm}[1]
  {\renewcommand\theinnercustomthm{#1}\innercustomthm}
  {\endinnercustomthm}

\newtheorem{innercustomprop}{Proposition}
\newenvironment{customprop}[1]
  {\renewcommand\theinnercustomprop{#1}\innercustomprop}
  {\endinnercustomprop}

\newtheorem{definition}{Definition}[section]
\newtheorem{prop}{Proposition}[section]
\newtheorem{theorem}{Theorem}[section]

\title{\LARGE \bf Diagrammatic Teaching of Orbitally Stable Systems
}

\author{Weiming Zhi$^{1,*}$ \and Tianyi Zhang$^{1}$ \and Matthew Johnson-Roberson$^{1}$
\thanks{$^{*}$email: {\tt\small wzhi@andrew.cmu.edu}.}%
\thanks{$^{1}$ Robotics Institute, Carnegie Mellon University, Pittsburgh, PA, USA}%
}

\maketitle

\begin{abstract}
Diagrammatic Teaching is a paradigm for robots to acquire novel skills, whereby the user provides 2D sketches over images of the scene to shape the robot's motion. In this work, we tackle the problem of teaching a robot to approach a surface and then follow cyclic motion on it, where the cycle of the motion can be arbitrarily specified by a \emph{single} user-provided sketch over an image from the robot's camera. Accordingly, we contribute the \emph{Stable Diffeomorphic Diagrammatic Teaching} (SDDT) framework. SDDT models the robot's motion as an \emph{Orbitally Asymptotically Stable} (O.A.S.) dynamical system that learns to stablize based on a single diagrammatic sketch provided by the user. This is achieved by applying a \emph{diffeomorphism}, i.e. a differentiable and invertible function, to morph a known O.A.S. system. The parameterised diffeomorphism is then optimised with respect to the Hausdorff distance between the limit cycle of our modelled system and the sketch, to produce the desired robot motion. We provide novel theoretical insight into the behaviour of the optimised system and also empirically evaluate SDDT, both in simulation and on a quadruped with a mounted 6-DOF manipulator. Results show that we can diagrammatically teach complex cyclic motion patterns with a high degree of accuracy.
\end{abstract}


\section{Introduction}
Specifying the desired behaviour for a robot has traditionally involved crafting a cost function and solving an optimisation problem. This process can often be complicated and require trial and error. Another way to generate desired robot motion has been Learning from Demonstration (LfD)~\cite{ravichandar2020recent}, where an expert demonstrates the movement to the robot. The demonstrations are often provided by \emph{Kinesthetic Teaching}, where the user physically handles the robot, or via teleoperation, which requires an additional remote controller. Both approaches face challenges when operating on robots with high degrees of freedom. Diagrammatic Teaching~\cite{zhi2023learning} is a recently introduced paradigm that circumvents physical contact and teleoperation, where the user specifies robot skills by sketching examples of the robot's end-effector motion on images of the scene. However, Diagrammatic teaching was restricted to learning distributions of motion, and it was unclear how to extract policies from the sketched. In this paper, we explore using user sketches as a medium for the user to shape the motion of the robot, and demonstrate the ability to construct orbitally stable policies from a single sketch on a camera image.

We focus on generating robot end-effector motion that approaches a flat surface and converges to a continuous periodic motion on it.  This motion pattern arises in many tasks, such as painting, wiping or sanding a surface. We represent the robot's end-effector motion as a dynamical system where trajectories of this system will eventually converge to the limit cycle. Dynamical systems with this convergence property are known to be \emph{Orbitally Asymptotically Stable} (O.A.S.). This paper aims to develop methodologies to learn robot policies, represented by O.A.S. dynamical systems, that are shaped by diagrammatic sketches provided by the user. Unlike LfD systems, generated motion trajectories do no follow a set of demonstrated motion trajectories, but are instead required to converge onto a limit cycle specified by a \emph{single} sketch. 

\begin{figure}[t]
\centering
\begin{subfigure}{.24\textwidth}
  \centering
  \fbox{\includegraphics[width=\linewidth]{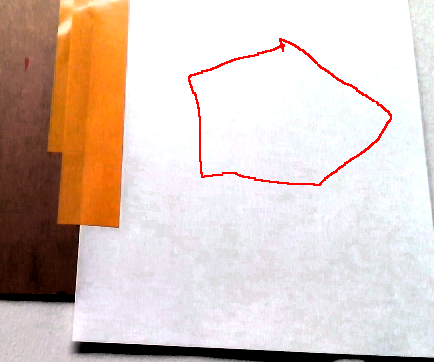}}  
\end{subfigure}%
\begin{subfigure}{.24\textwidth}
  \centering
  \fbox{\includegraphics[width=\linewidth]{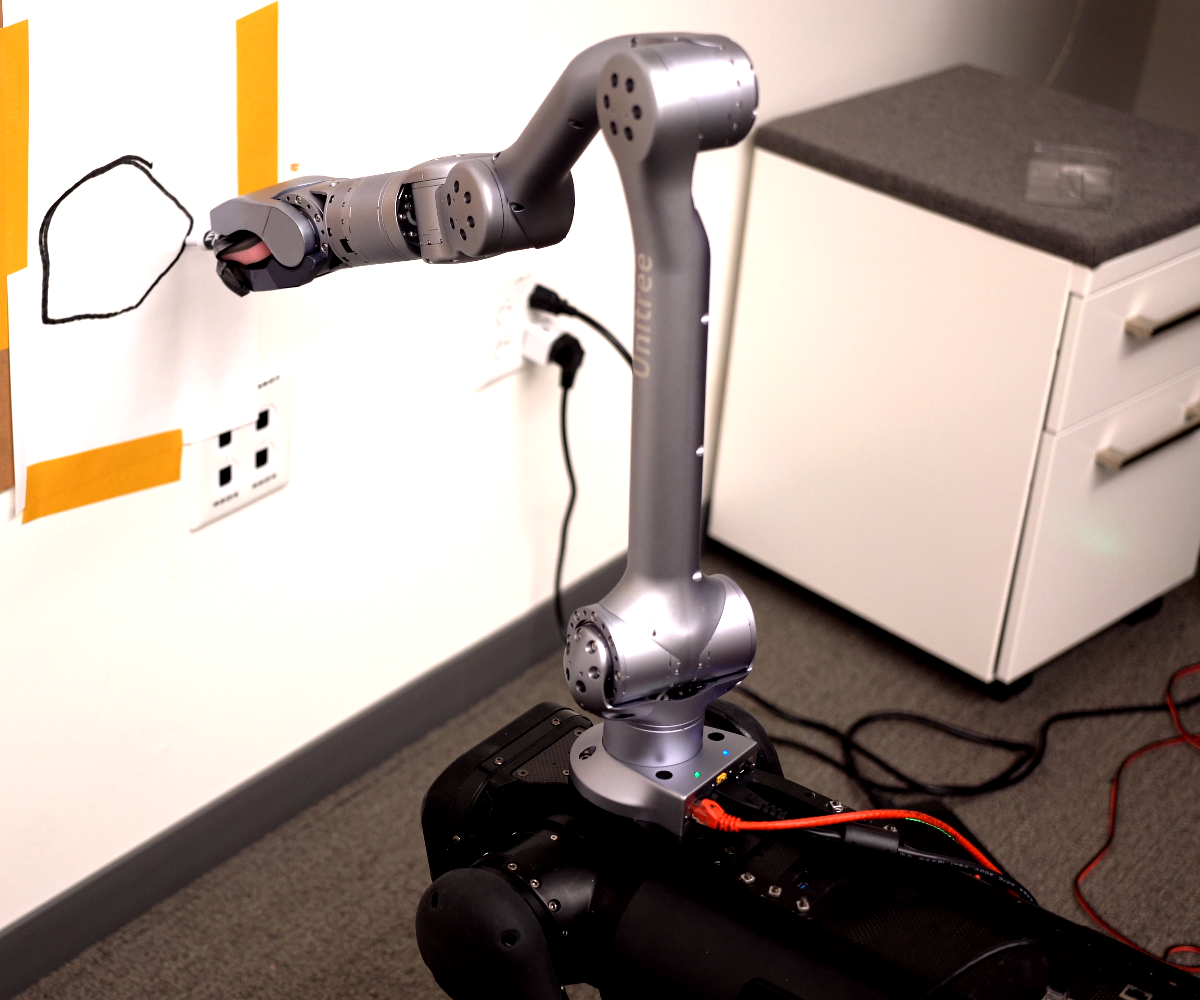}}  
\end{subfigure}%
\caption{Diagrammatic teaching is a paradigm to interface with robots by drawing sketches over camera images. We contribute SDDT to diagrammatically teach robots robot policies that approach a surface in view and stabilise at cyclic motions of the provided shape on the surface. (Left) A sketch of the desired pentagon-shaped cycle (in red) is provided by the user from the egocentric view of the robot. (Right) The resulting policy forces the end-effector to quickly approach the surface, and then stabilise to continuously trace out the shape of the provided sketch.}
\label{fig:1}
\end{figure}

We introduce \emph{Stable Diffeomorphic Diagrammatic Teaching} (SDDT), a novel framework for diagrammatic teaching to learn policies that produce periodic and surface-approaching motions. SDDT allows the user to provide input by sketching the shape of the desired limit cycle onto an image of the surface. We use the insight that when two dynamical systems map to one another via a \emph{diffeomorphism} (a differentiable and invertible function), these systems have the same stability properties~\cite{Diff_templates,Fast_diff_int}. SDDT learns O.A.S. systems by optimising a parameterised diffeomorphism to ``morph'' a system that is known to be O.A.S., such that the limit cycle matches the desired shape. We develop a novel Hausdorff distance-based loss for the corresponding optimisation. 

Then, we provide both theoretical guarantees into what classes of shapes the limit cycle can be ``morphed'' into. We complement our theoretical insights with empirical evidences of the efficacy of our proposed framework. These include real-world experiments on a quadruped with a mounted manipulator. SDDT is particularly appealing for mobile manipulators (like quadrupeds with installed arms \cref{fig:1}), where an egocentric view, readily available from the onboard vision system, is used to prompt diagrammatic sketches from the user. 

The remainder of the paper is organised as follows: we begin by reviewing related work in \cref{sec:related_work}, and then provide some background knowledge to understand SDDT in \cref{sec:preliminaries}. We introduce the methodology of SDDT and provide theoretical results in \cref{sec:SDDT}, followed by empirical results in \cref{sec:experiments}. We end the paper with conclusions and future directions in \cref{sec:conclusion}.

\section{Related Work}\label{sec:related_work}
\subsection{Robot Motion Generation and Diagrammatic Teaching}
Generating robot motion is a central problem in robotics. This can be done in a motion planning fashion~\cite{Motion_planning}, where an optimisation problem needs to be constructed and a motion planner~\cite{rrts,CHOMP,RMPs} is then used to find a solution to the problem. Motion planning approaches are beneficial in that once the constraints and costs for the motion have been specified, the task of motion generation is primarily ``off-loaded'' to the planner, and the solution inherits theoretical guarantees, such as probabilistic completeness~\cite{rrtstar}. However, many natural motion patterns cannot be easily distilled into a simple cost function, and additionally, the construction of the optimisation problem requires technical expertise. Another approach for specifying robot motion has been Learning from Demonstration (LfD)~\cite{ravichandar2020recent}, where a human expert physically handles the robot to trace out the desired motion (i.e. kinesthetic teaching)~\cite{learning_mps, promp} or teleoperation via specialised remote controllers. To allow demonstrations to be more easily and intuitively collected, \emph{Diagrammatic Teaching}~\cite{zhi2023learning} has been introduced as an alternative interface for the user to specify movement patterns to robots: the user is provided with images and prompted to provide sketches, which are subsequently used to construct a model of robot motions. This work falls within the Diagrammatic Teaching paradigm, where the robot's motion is shaped by a sketch from the user, and not by demonstrations of trajectories.    

\subsection{Dynamical Systems as Robot Policies}
In many LfD and motion generation problem formulations, robot policies are modelled by state-dependent dynamical systems~\cite{geofabs, GeoFab_gloabL_opt}. Enforcing the convergence properties of dynamical systems has been shown to increase the robustness of the robot policy and enables prior knowledge to be imbued into the system~\cite{stable_vect_field_example}. However, previous methods exclusively focus on systems that converge to a single fixed point, instead of an orbit. Additionally, these systems are learned in a LfD setup where the training data is an entire set of collected expert trajectories, in the form of sequences of end-effector or joint positions. Examples of such methods include~\cite{Ijspeert2013DynamicalMP,geofabs,Saveriano2020AnEA,dynamic_p,Euclideanising,Diff_templates}, In particular, methods~\cite{Diff_templates, Euclideanising} also take a diffeomorphic learning approach, but only considered stability with respect to convergence to a fixed point. Our work is unique from these previous approaches, in that we study learning stable systems within the diagrammatic teaching paradigm. Our approach does not require multiple kinesthetic demonstrations, and instead simply requires a single sketch provided by the user.

\section{Preliminaries}\label{sec:preliminaries}
Here, we introduce the necessary background concepts for the presentation of SDDT in \cref{sec:SDDT}.

\subsection{Robot Motion Generation via Dynamical Systems}
In this work, we shall directly model the robot's end-effector position $\bb{x}\in\mathbb{R}^{3}$. We represent the robot's policy as a first-order time-invariant dynamical system.
\begin{align}
\bb{\dot{x}}(t)=f(\bb{x}(t)), && \bb{\dot{x}}(0)=\bb{x}_{0},\label{eqn:system}
\end{align}
where $f:\mathbb{R}^{3}\rightarrow\mathbb{R}^{3}$ is a non-linear mapping between $\bb{x}$ and its time derivative $\bb{\dot{x}}$, and $\bb{x}_{0}$ is the initial condition. Individual motion trajectories $\xi$ of time duration $t\in\mathbb{R}$ can be obtained via integration, 
\begin{equation}
\xi(t,\bb{x}_{0})=\bb{x}_{0}+\int_{0}^{t}\bb{\dot{x}}(s)\mathrm{s},
\end{equation}
where the integral can be evaluated using a numerical ODE integrator, such as Euler's method. Modelling the robot's policy, rather than individual trajectories, has the benefit of being robust to perturbations. That is, at any end-effector state after perturbation, the robot can follow the dynamical system and does not track a pre-determined trajectory. In our problem setup, we may wish to additionally constrain the end-effector to be orthogonal to the surface, fixing its rotation.

\subsection{O.A.S. Systems}
We are interested in understanding the long-term behaviour of the dynamical system, namely, what happens to the trajectories after a long integration duration. Will the solution eventually converge to fixed points, a limit cycle, or diverge and blow up? We are interested in robot motion which approaches a surface and converges onto a cyclic motion on that surface. This requires the dynamical system to be \emph{Orbitally Asymptotically Stable (O.A.S.)} with a limit cycle. Here, we give a definition for O.A.S.

\begin{definition} [O.A.S. Stability]
A dynamical system $\bb{x}=f(\bb{x})$ is Orbitally Asymptotically Stable (O.A.S.) if for any initial condition $\bb{x}_{0}$ within a region of state-space $\mathcal{X}$, we have
\begin{align}
\lim_{t\rightarrow\inf}\min_{\tau\in[0,T]}\lvert\lvert \bb{x}(t)-\bb{\bar{x}(\tau)}\lvert\lvert=0,
\end{align}
where $\bb{\bar{x}}(\tau)$ is a solution of the system. Furthermore, $\bb{\bar{x}}(\tau)$ is periodic, i.e. $\bb{\bar{x}}(\tau)=\bb{\bar{x}}(\tau+T)$, where $T\in\mathbb{R}^{+}$ is the period of the cycle. Here, $\bb{\bar{x}}(\tau)$ is known as a ``limit cycle'' and $\mathcal{X}$ is known as the ``basin of attraction''.
\end{definition}

\subsection{Invertible Neural Networks}
\emph{Diffeomorphisms}, which are differentiable and invertible functions, are crucial building blocks of our proposed framework. Diffeomorphisms can be parameterised by Invertible Neural Networks (INNs). INNs are function approximators that are invertible by definition and have easily computable Jacobians~\cite{Freia}. We use \emph{Coupling-based INNs}~\cite{Real_nvp} which contain the reversible block introduced, where the split the input $\bb{u}$ into halves, $\bb{u} = [\bb{u}_{1},\bb{u}_{2}]$, and the output $\bb{v}$ into halves, $\bb{v}=[\bb{v}_{1},\bb{v}_{2}]$. We learn four fully-connected neural networks, $p_{1}, p_{2}, q_{1}, q_{2}$ such that,
\begin{align}
\bb{v}_{1}&=\bb{u}_{1}\odot\exp(p_{2}(\bb{u}_{2}))+q_{2}(\bb{u}_{2}), \\ \bb{v}_{2}&=\bb{u}_{2}\odot\exp(p_{1}(\bb{v}_{1}))+q_{1}(\bb{v}_{2}),
\end{align}
where $\odot$ denotes the Hadamard product. By construction, the inverse is given as follows:
\begin{align}
\bb{u}_{1}&=(\bb{v}_{1}-q_{2}(\bb{u}_{2}))\odot\exp(p_{2}(\bb{u}_{2})), \\ \bb{u}_{2}&=(\bb{v}_{2}-q_{1}(\bb{v}_{1}))\odot\exp(p_{1}(\bb{v}_{1})).
\end{align}
As such, the INN is able to enforce invertibility without the functions $p_{1}, p_{2}, q_{1}, q_{2}$ being invertible. 
\section{Stable Diffeomorphic Diagrammatic Teaching}\label{sec:SDDT}
Stable Diffeomorphic Diagrammatic Teaching (SDDT) first presents the user with an image of the contact surface and prompts the user to sketch a closed shape on the image. The corresponding points in the robot's task space are found via ray-tracing the sketch onto the surface. We then minimise the distance between the limit cycle of a parameterised O.A.S. system and the set of corresponding points. 

This section is organised as follows: We shall first elaborate on how to construct a parameterised 3D dynamical system which is O.A.S. with a stable orbit on a surface (\cref{subsec:parameterise_goas}). We describe how to shape the system to match a sketch provided by the user (\cref{subsec:shaping}). Then, we provide theoretical guarantees that our model is sufficiently flexible to model any limit cycle that is smooth and closed (\cref{sec:guarentees}).

\subsection{Parameterising O.A.S. Systems via Diffeomorphisms}\label{subsec:parameterise_goas}

\begin{figure}[b]
\centering
\begin{subfigure}{.2\textwidth}
  \centering
  \includegraphics[width=\linewidth]{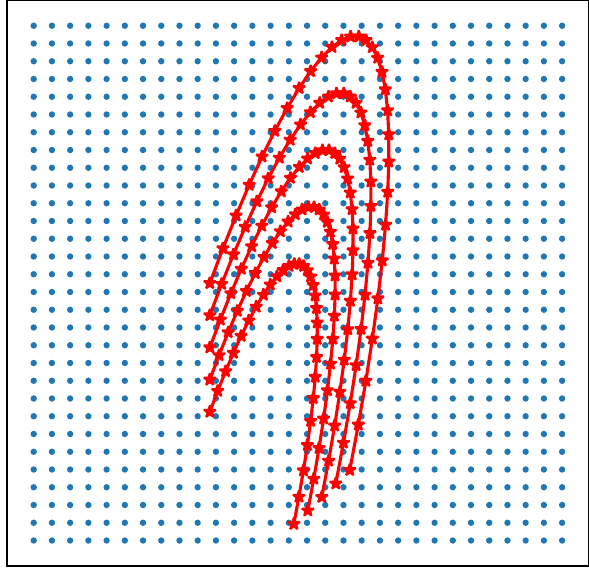}  
\end{subfigure}%
\begin{subfigure}{.2\textwidth}
  \centering
  \includegraphics[width=\linewidth]{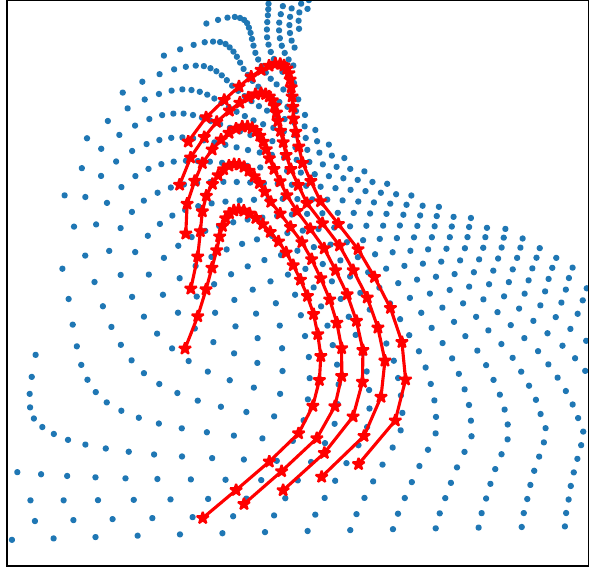}  
\end{subfigure}%
\caption{Diffeomorphisms can be thought of as ``morphing'' a dynamical system into one another. (Left) Five trajectories (red) of overlaid on grid points (blue); (Right) Morphed trajectories and the corresponding grid.}
\label{fig:morph}
\end{figure}

We can learn a desired O.A.S. system $\bb{\dot{x}}=f(\bb{x})$, by starting with a hand-designed \emph{base} O.A.S. system $\bb{\dot{y}}=g(\bb{y})$. We then learn a diffeomorphism $F$ such that $\bb{x}=F(\bb{y})$. Intuitively, we can think of the diffeomorphism to be ``morphing'' the base system into the desired system. A simple illustration of this is provided in \cref{fig:morph}. Throughout this section, we denote the state variables of the base system as $\bb{y}$, and that of the desired system as $\bb{x}$.

\begin{wrapfigure}{l}{0.28\textwidth}
  \centering
  \includegraphics[width=0.13\textwidth]{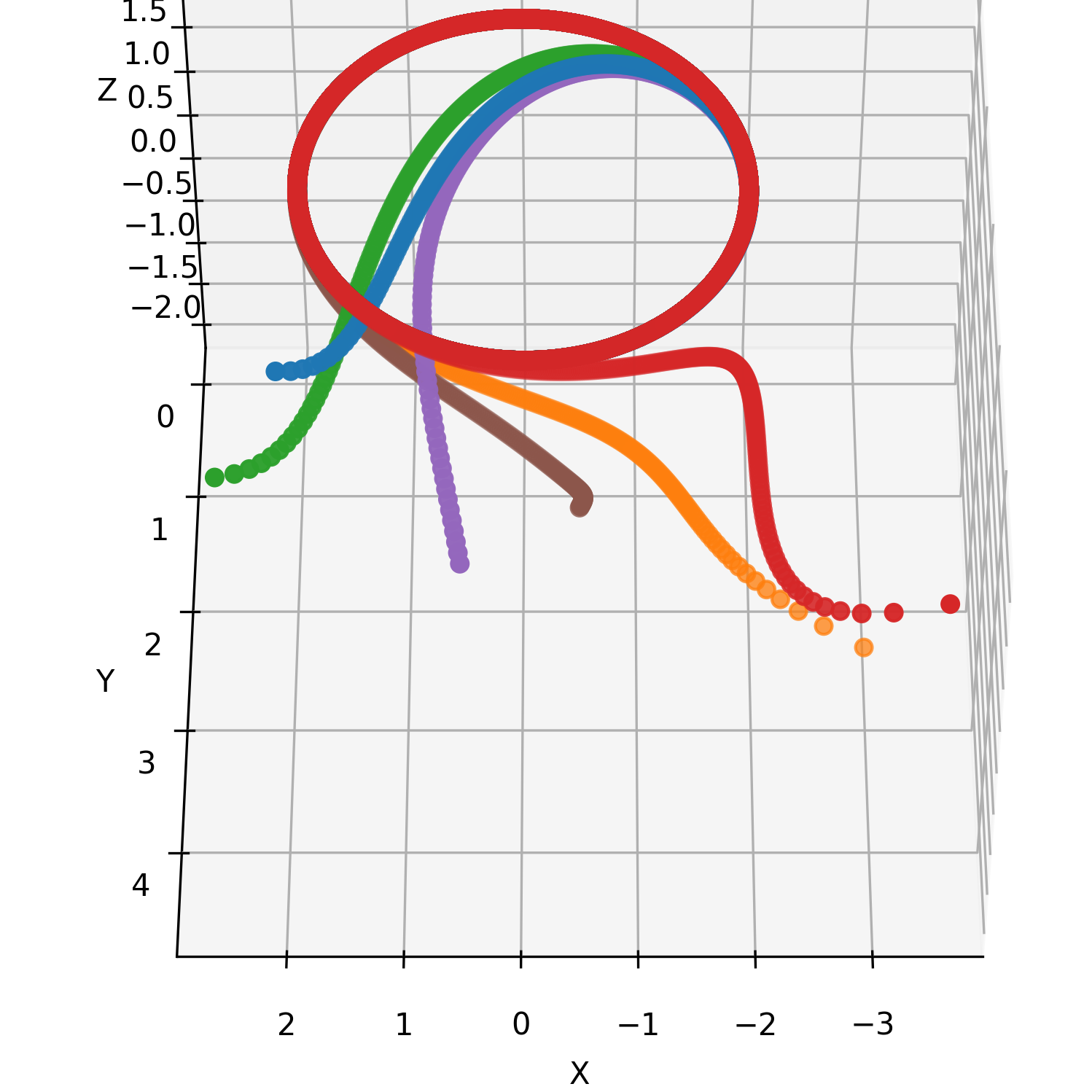}
  \includegraphics[width=0.13\textwidth]{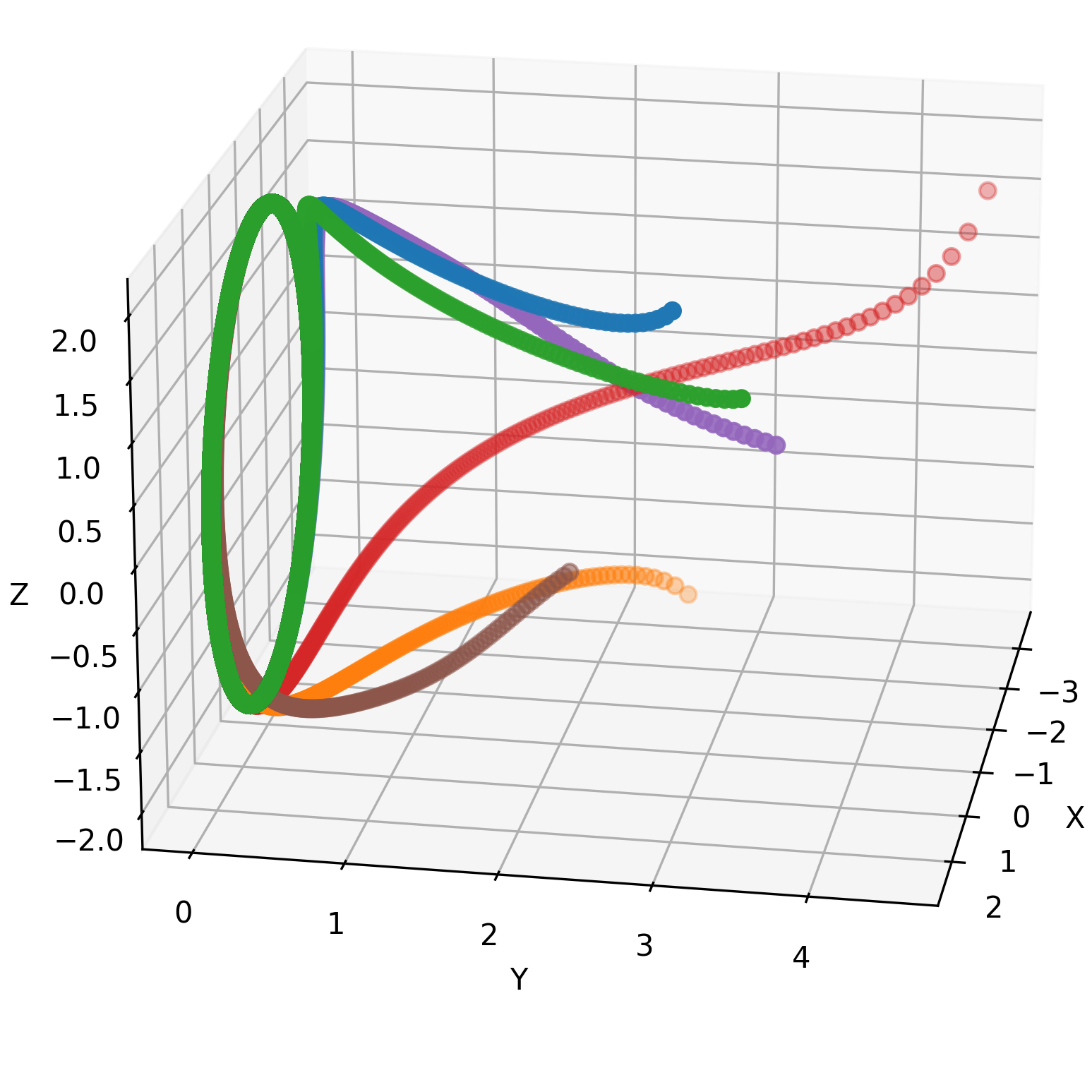}
  \caption{\small Trajectories converge to a limit cycle at $y=0$.}\label{fig:example_trajs}
\end{wrapfigure}

In this paper, we will by convention define the flat surface as the $x,z$-plane at $y=0$. We begin by constructing a simple base system to have a stable circular orbit on the surface. Consider a system where a polar coordinate system (with polar variables $r$ and $\omega$) is defined in the $x,z$-plane, with an additional attractor in the $y$-axis:

\begin{align}
\dot{r}=\mu(1-\frac{r^{2}}{R^{2}})r, && \dot{\omega}=1, && \dot{y}=-\alpha y,
\end{align}
where $\mu>0$ and $\alpha>0$ are parameters which control how fast the system converges. Trajectories of this system will converge to an equilibrium at $r=R$, $y=0$, and any $\omega$, for all $r\geq 0$. This system is O.A.S., with a basin of attraction $\mathcal{X}=\{(r,\omega,y)\lvert r>0, \omega,y\in\mathbb{R}\}$. Example trajectories of this system are shown in \cref{fig:example_trajs}. We can transform the coordinates into Cartesian coordinates as the system:

\begin{align}
\bb{\dot{y}}=g(\bb{y})=\begin{bmatrix}
\dot{x}\\
\dot{y}\\
\dot{z}
\end{bmatrix}= 
\begin{bmatrix}
-z + \mu \left(1 - \frac{x^2+z^2}{R^2}\right) x, \\
-\alpha y, \\
x + \mu \left(1 - \frac{x^2+z^2}{R^2}\right) z,
\end{bmatrix}.\label{eqn:base_sys}
\end{align}
which has a limit cycle $L:=\{(x,y,z)\in\mathbb{R}^{3}|x^2+z^2=R^2, y=0\}$.

Dynamical systems with state variables satisfying $\bb{x}=F(\bb{y})$, where $F$ is a diffeomorphism, are topologically conjugates of one another. They can be thought of as the same system under a change of coordinate systems and their stability characteristics are not altered (proof in~\cite{Fast_diff_int,Euclideanising, manifold_textbook}).
We seek a mapping $F$, such that no change is made on the $y$ axis while shaping the circle on the $x,z$ plane $x^{2}+z^{2}=R^2$ to match the provided data. Therefore, we decompose $F$ into separate functions, using an INN to learn a diffeomorphism on the $x,z$ axes and leaving the $y$-axis with the identity function. Specifically,  
\begin{align}
\!\bb{x}\!=\!F\!(\bb{y})\!,&&\!\text{where }[x_{x}, x_{z}]\!=\!\mathrm{INN}_{\bm{\theta}}([y_{x},y_{z}])\!,&&\!x_{y}\!=\!y_{y},\label{eqn:diff}
\end{align}
where $\bb{x}=[x_{x},x_{y},x_{z}]$, $\bb{y}=[y_{x},y_{y},y_{z}]$ and $INN_{\bm{\theta}}$ is an invertible neural network with parameters $\bm{\theta}$. 

The desired O.A.S. system dynamics $\bb{\dot{x}}=f(\bb{x})$ is now related to that of the base O.A.S. system $\bb{\dot{y}}=g(\bb{y})$ via the chain rule:
\begin{align}
\bb{\dot{x}}=f(\bb{x})=J_{F}(F^{-1}(\bb{x}))g(F^{-1}(\bb{x})),\label{eqn:desired_sys}
\end{align}
where $J_{F}(F^{-1}(\bb{x}))$ is the Jacobian of $F$ at $F^{-1}(\bb{x})$.

\subsection{Learning the Parameterised System via the Hausdorff Distance}\label{subsec:shaping}
This section elaborates on how to train $F$ defined in \cref{eqn:diff} such that the limit cycle of $\bb{\dot{x}}=f(\bb{x})$ matches the user's sketch. This involves projecting the user's sketch to the surface via ray-tracing. Then, defining and minimising a loss between the set of projected points on the surface to the limit cycle of our dynamical system model.

\begin{figure}[t]
\centering
\begin{subfigure}{.22\textwidth}
  \centering
  \includegraphics[width=\textwidth]{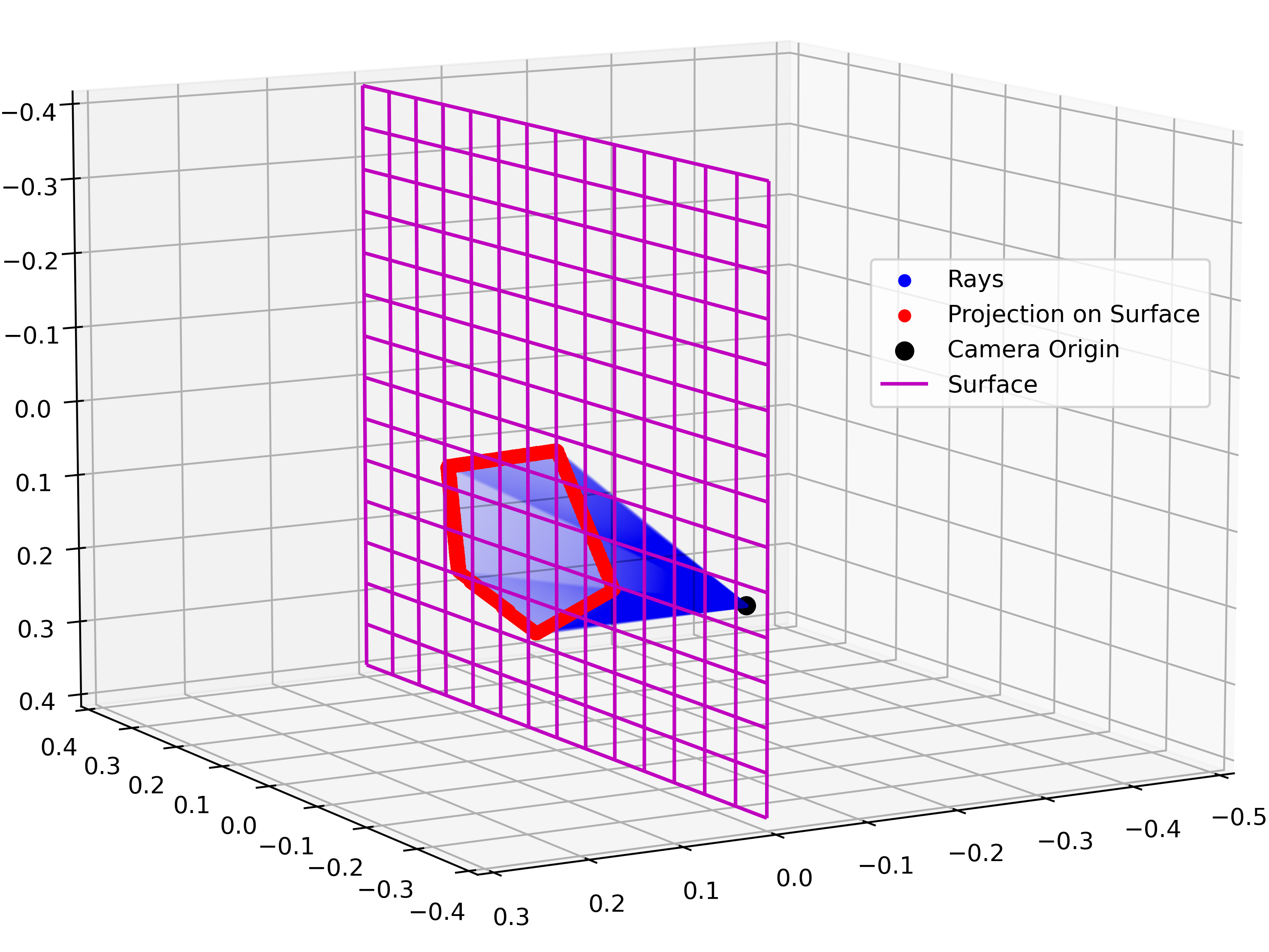}
  \caption{\small Visualisation of ray-tracing (rays in blue) a pentagon shape (in red) onto a surface (magenta). }\label{fig:example_ray_trace}
\end{subfigure}%
\hspace{1em}
\begin{subfigure}{.22\textwidth}
  \centering
  \includegraphics[width=\textwidth]{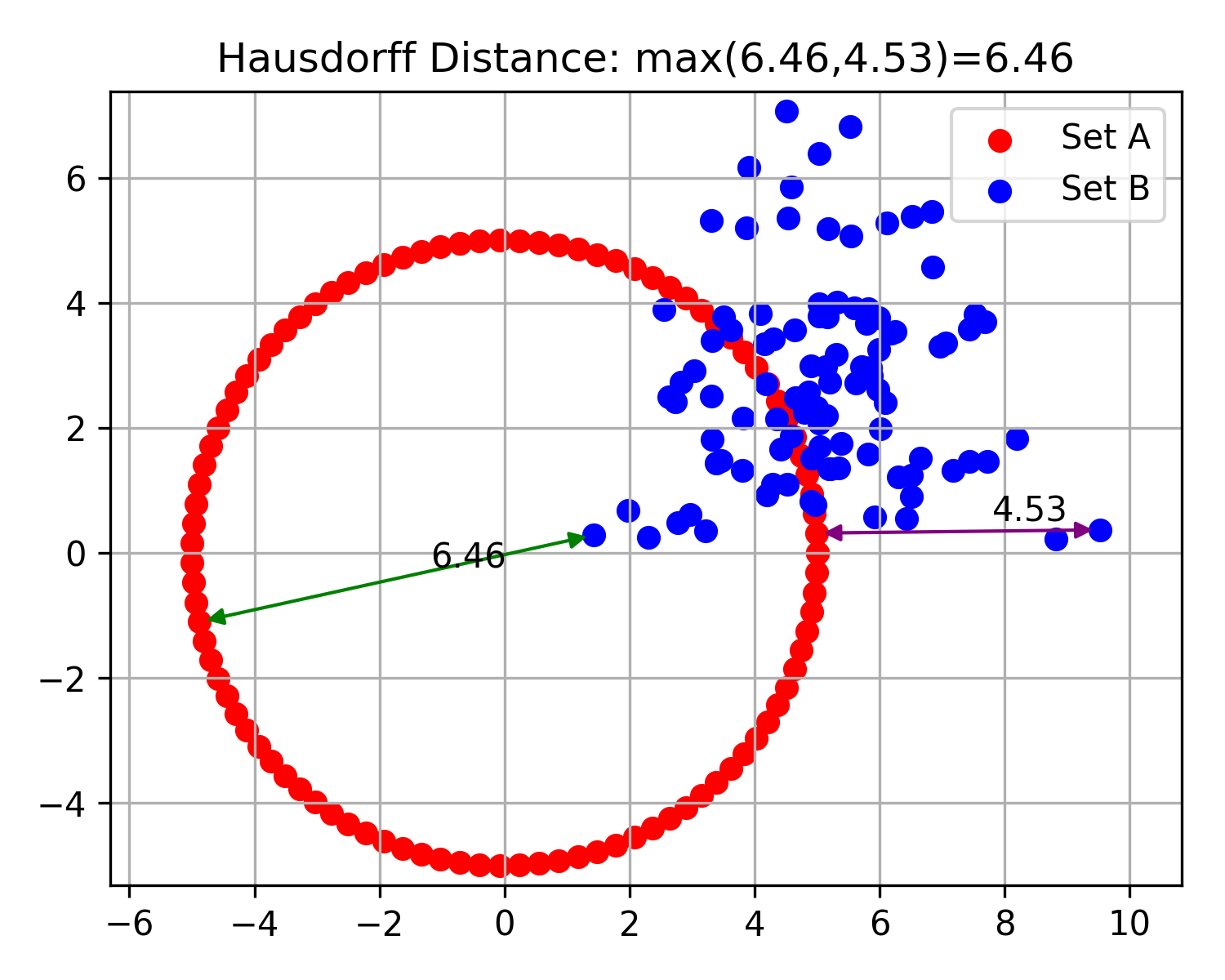}
  \caption{\small Example contributing distances (green, purple) of the Hausdorff distance between two (red and blue) point sets.}\label{fig:example_HD}
\end{subfigure}%
\end{figure}

\begin{figure}[t]
\setlength{\fboxsep}{0pt}
\centering
  \centering
\fbox{\includegraphics[width=.48\linewidth]{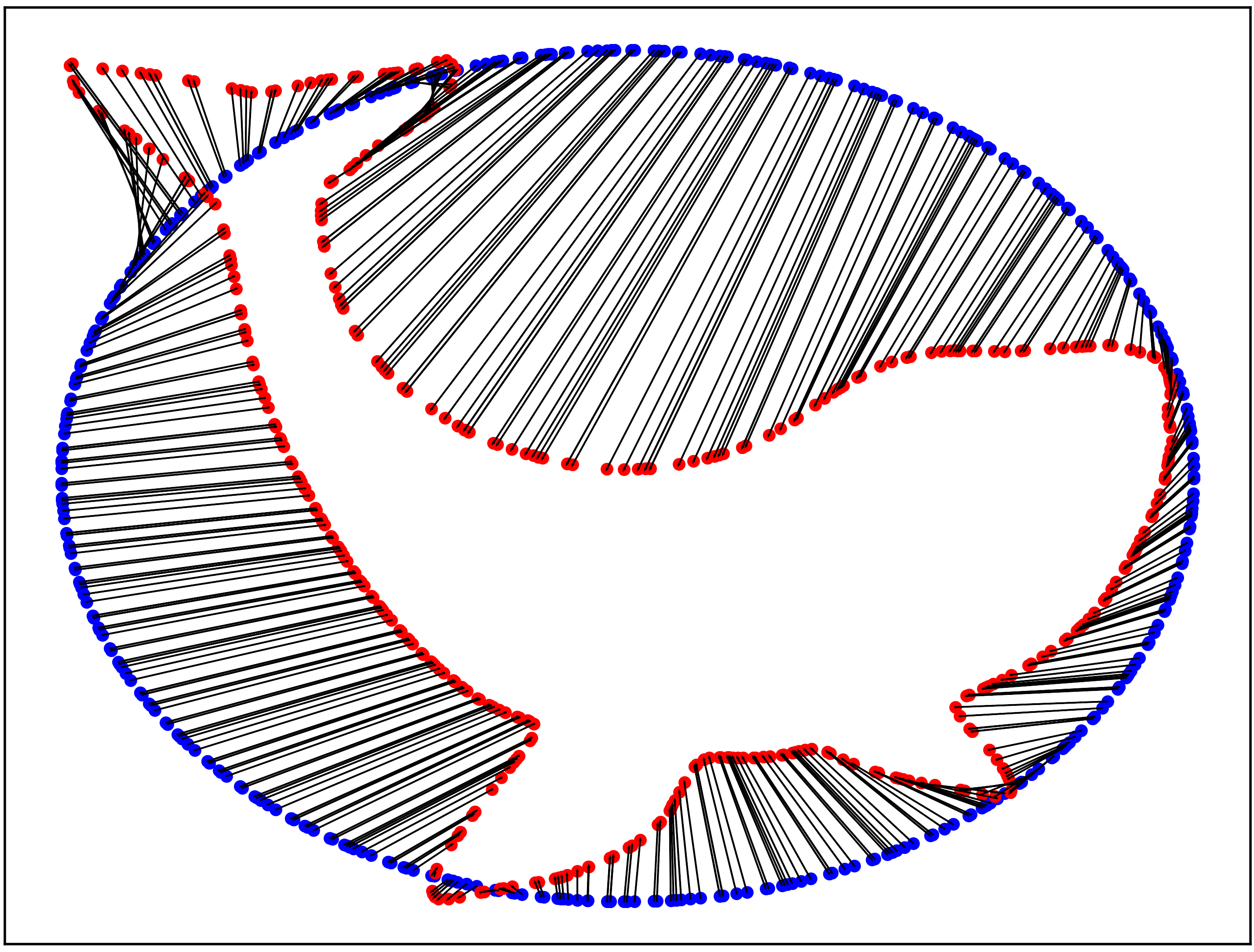}}%
\hspace{0.1em}
\fbox{\includegraphics[width=.48\linewidth]{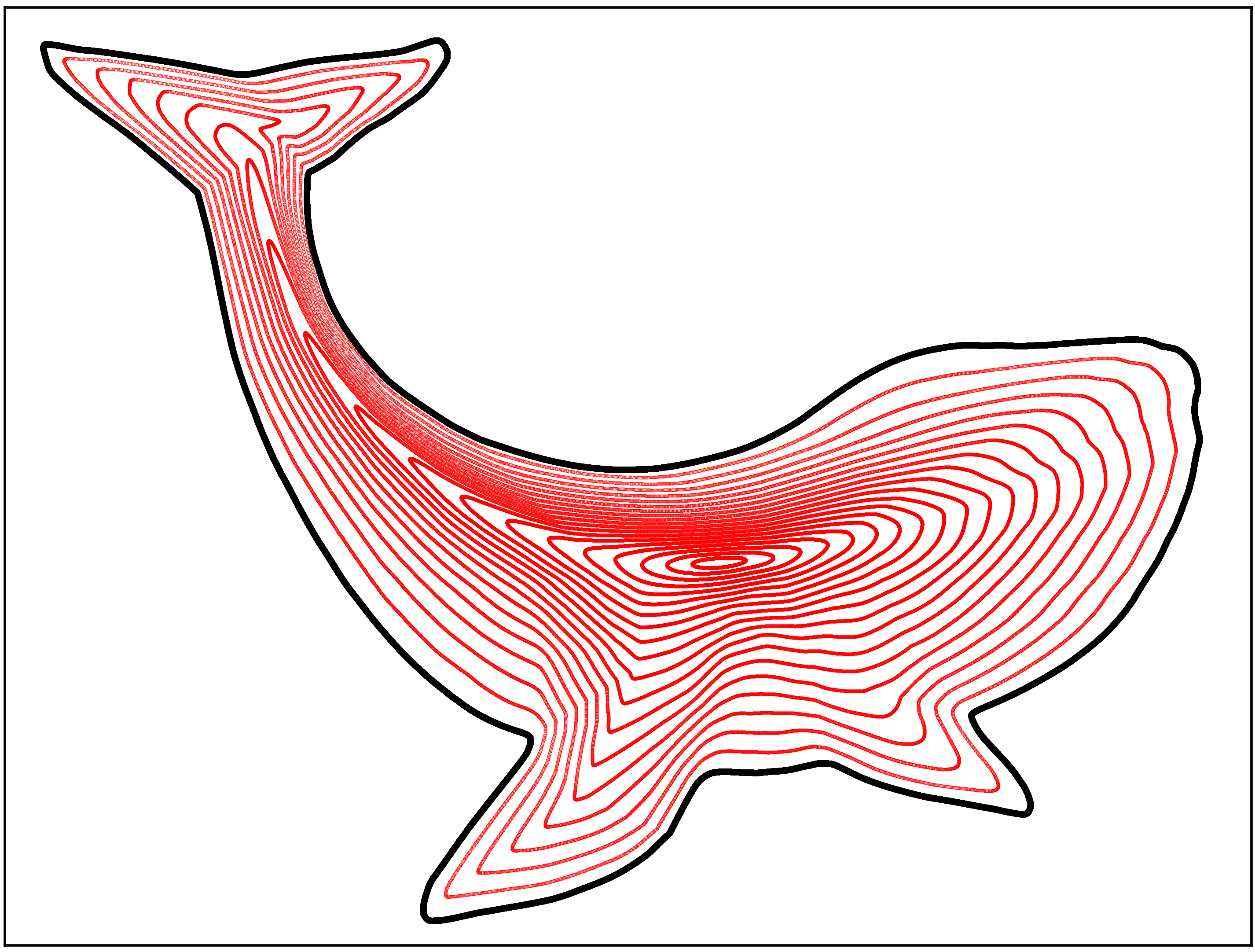}}%
\caption{We visualise how the ambient space is morphed by the learnt diffeomorphism: (Left) Transport map with points on the base system (blue) mapped (shown by grey line) onto those of the learned system (red); (Right) Concentric circles passed through the diffeomorphism to match the desired shape.}
\label{fig:analysis}
\end{figure}

\begin{figure*}[t]
\centering
\begin{subfigure}{.495\textwidth}
  \centering

  \fbox{
  \includegraphics[width=.33\linewidth]{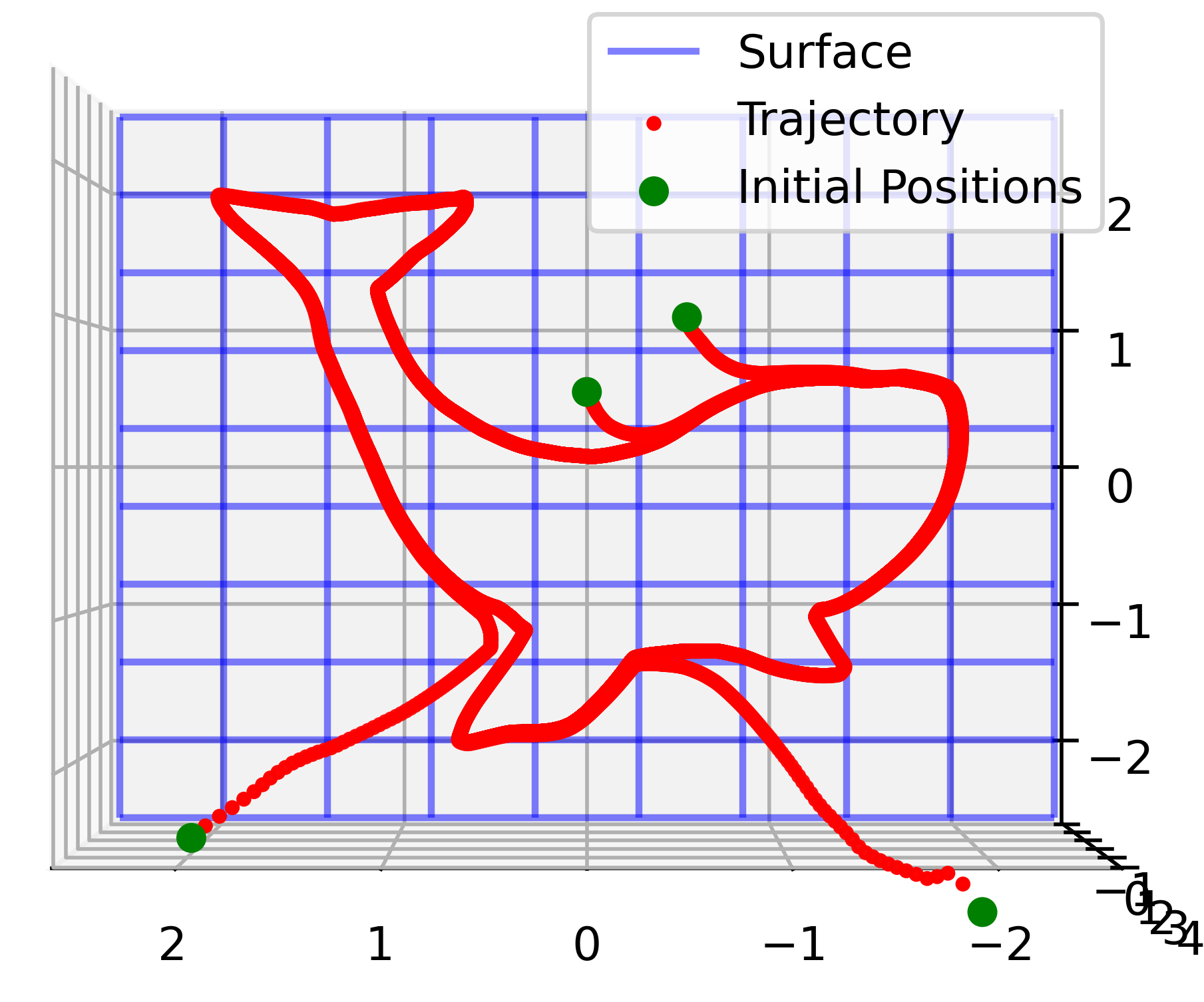}%
  \includegraphics[width=.33\linewidth]{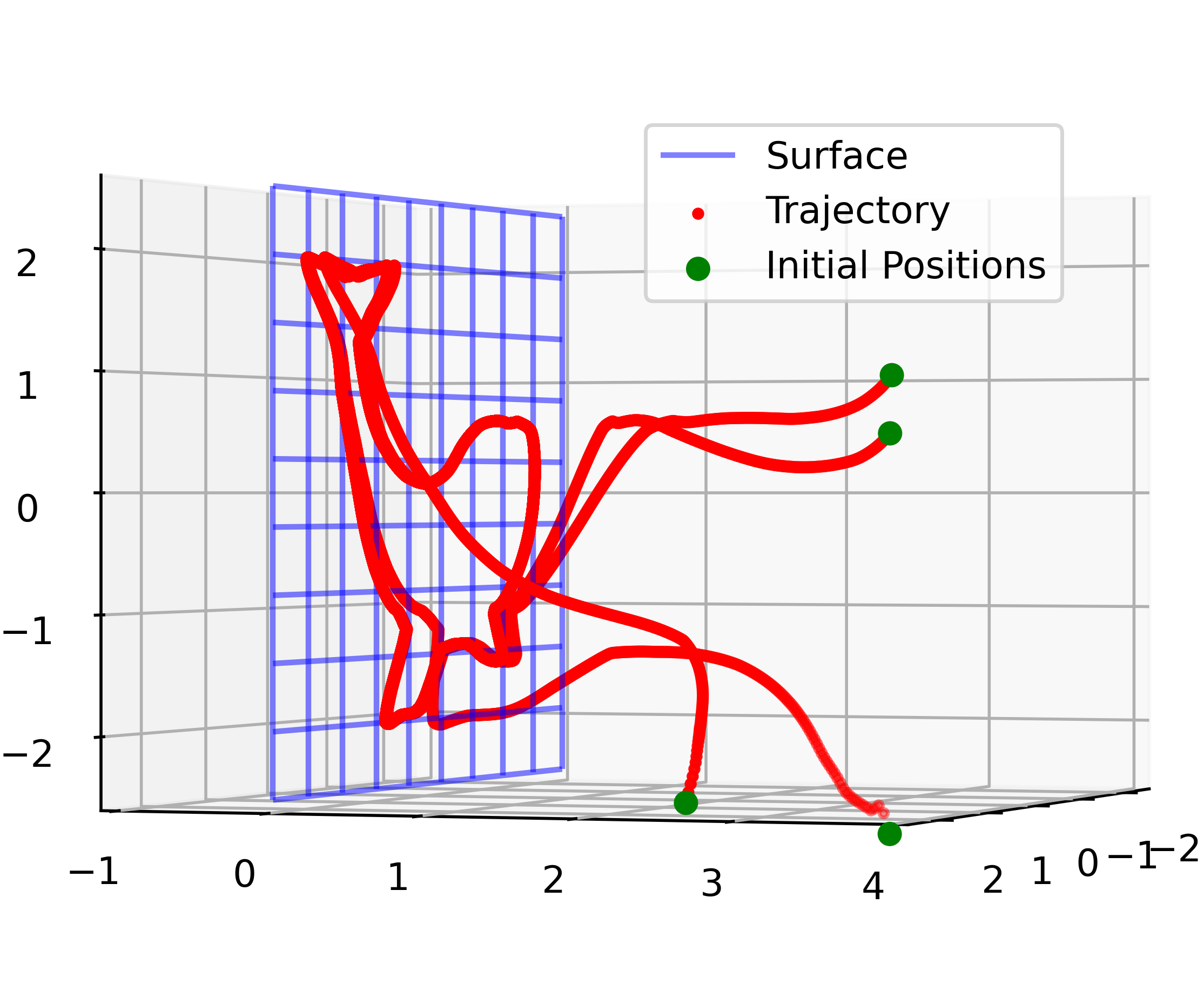}%
  \includegraphics[width=.33\linewidth]{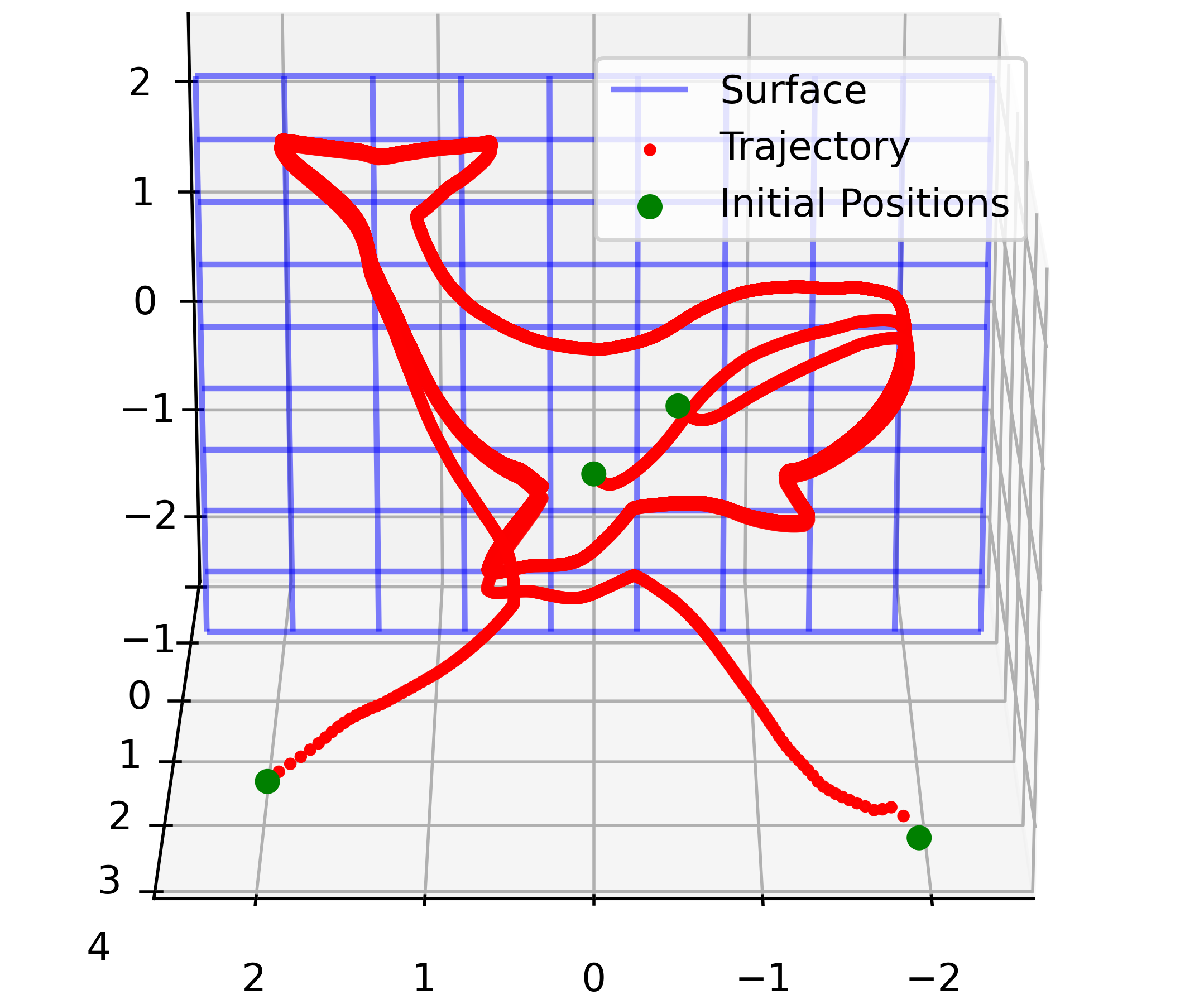}}  
\end{subfigure}%
\begin{subfigure}{.495\textwidth}
  \centering
  \fbox{
  \includegraphics[width=.33\linewidth]{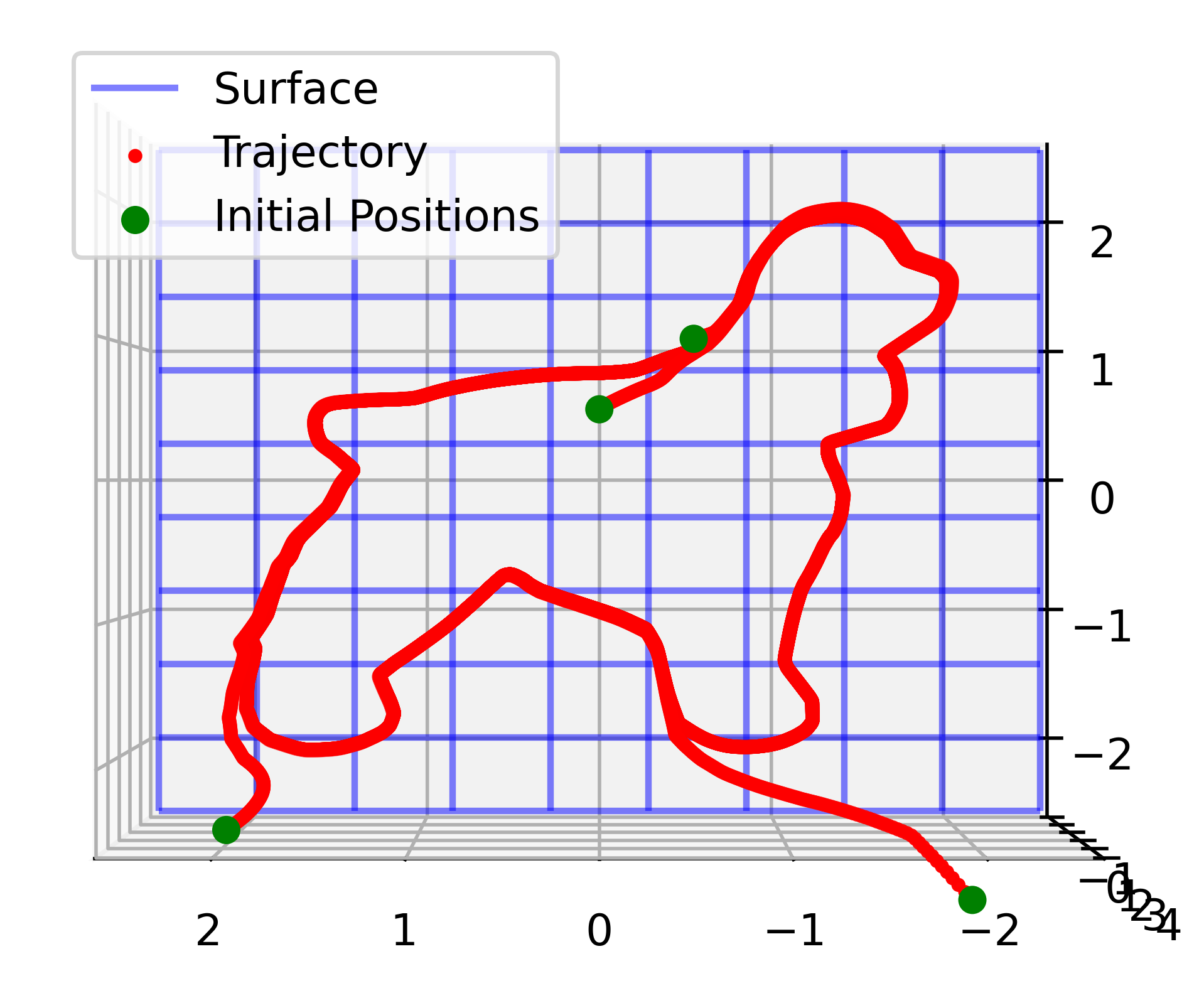}%
  \includegraphics[width=.33\linewidth]{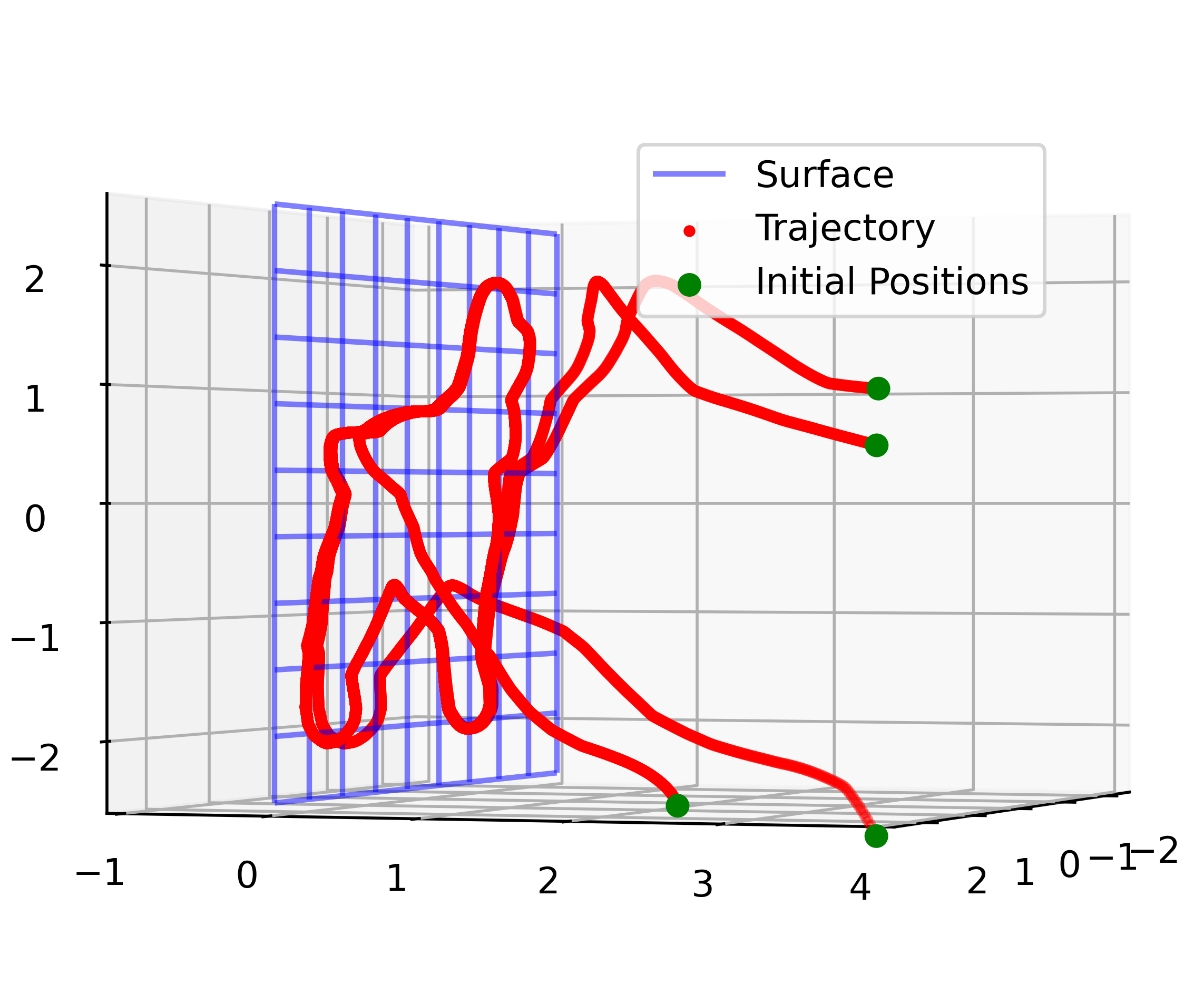}%
  \includegraphics[width=.33\linewidth]{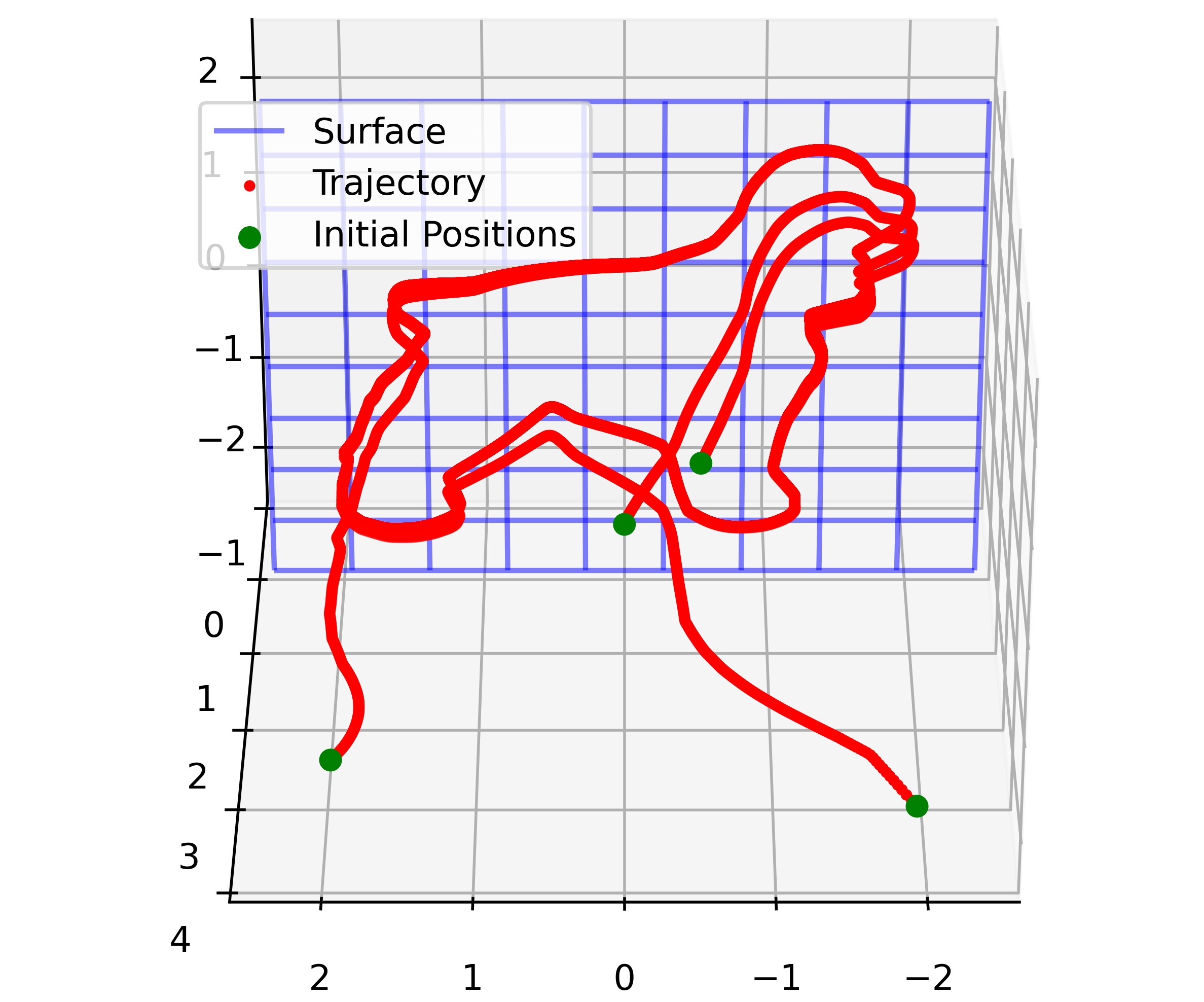}}  
\end{subfigure}%

\begin{subfigure}{.495\textwidth}
  \centering
  \fbox{
  \includegraphics[width=.33\linewidth]{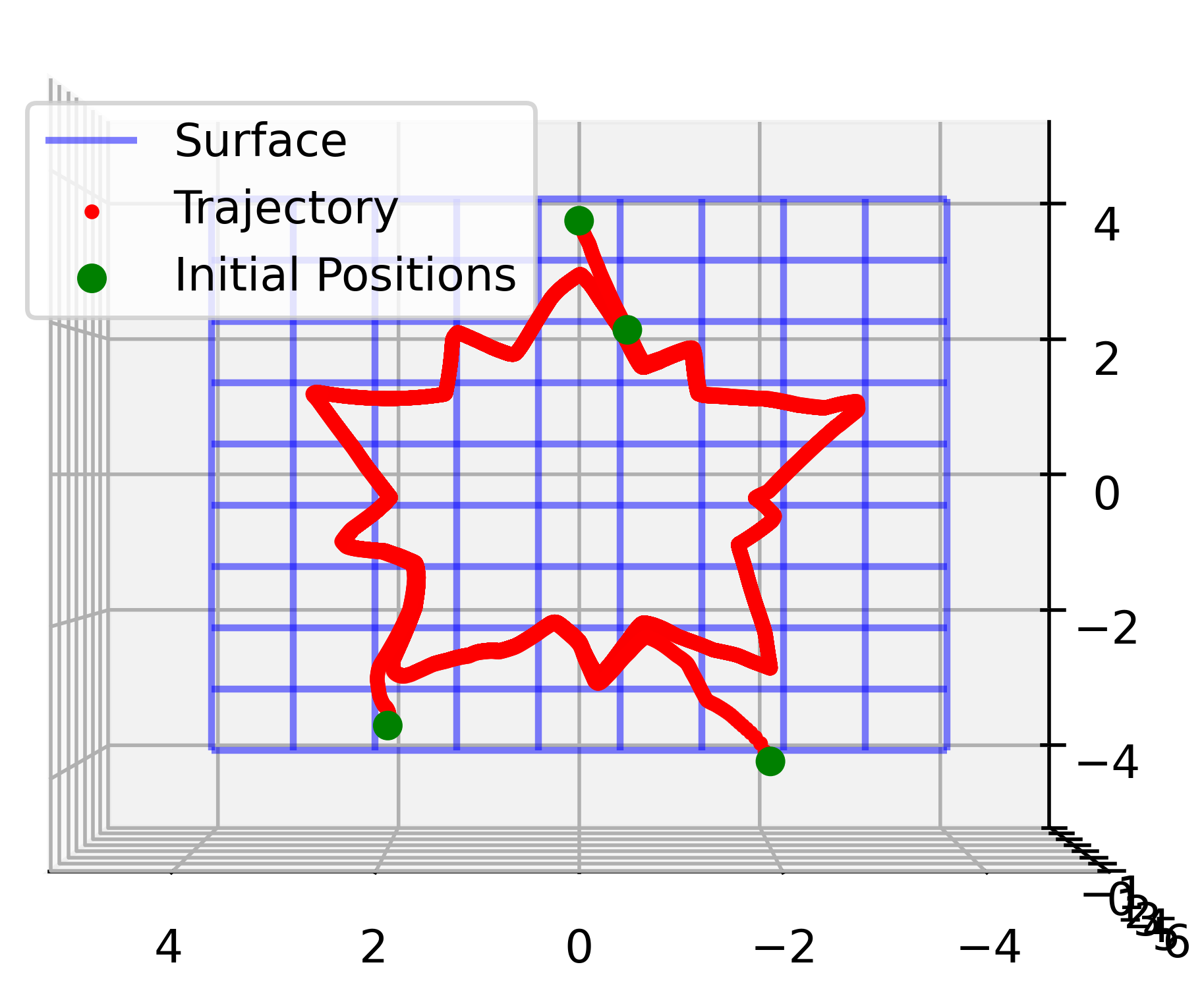}%
  \includegraphics[width=.33\linewidth]{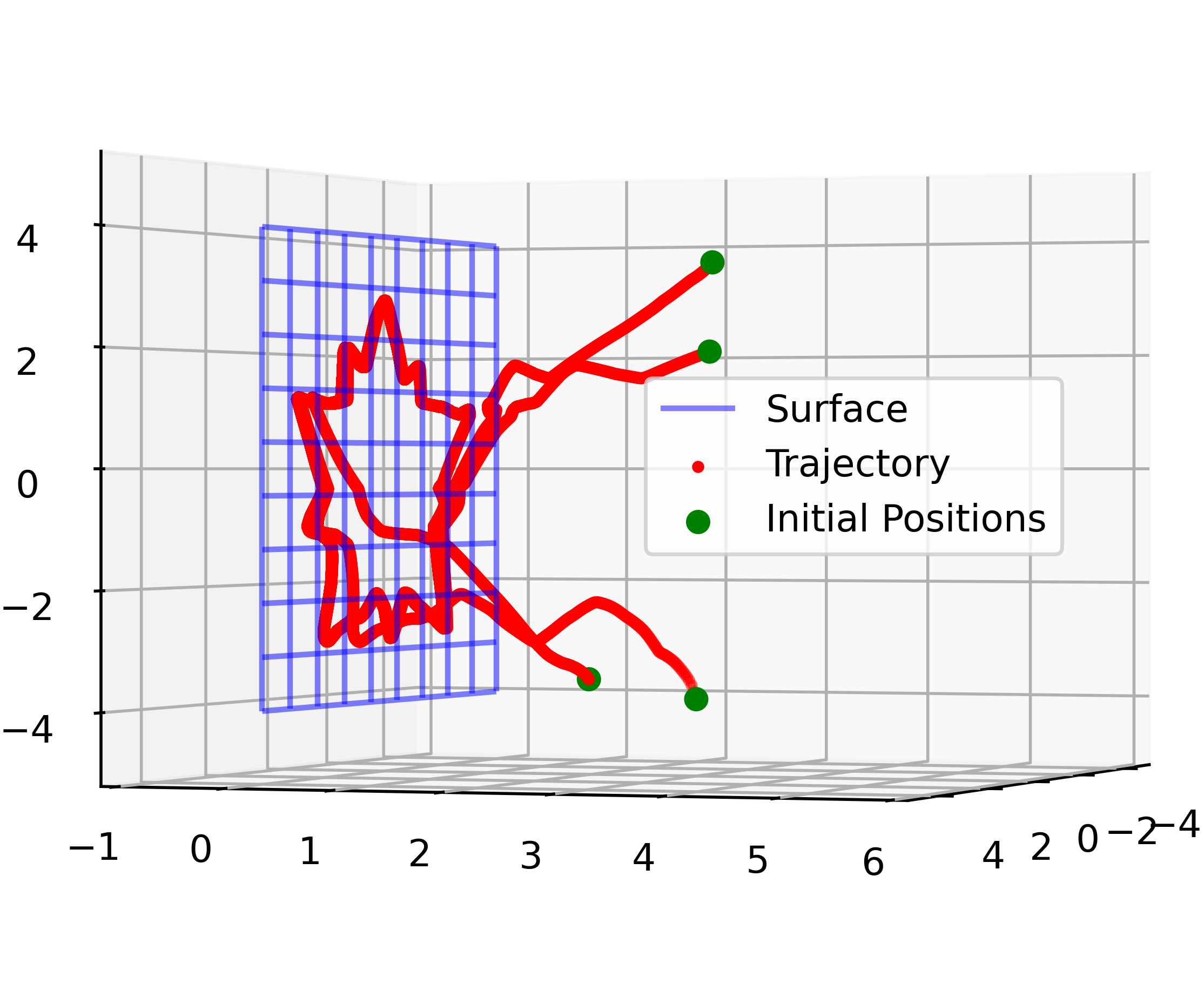}%
  \includegraphics[width=.33\linewidth]{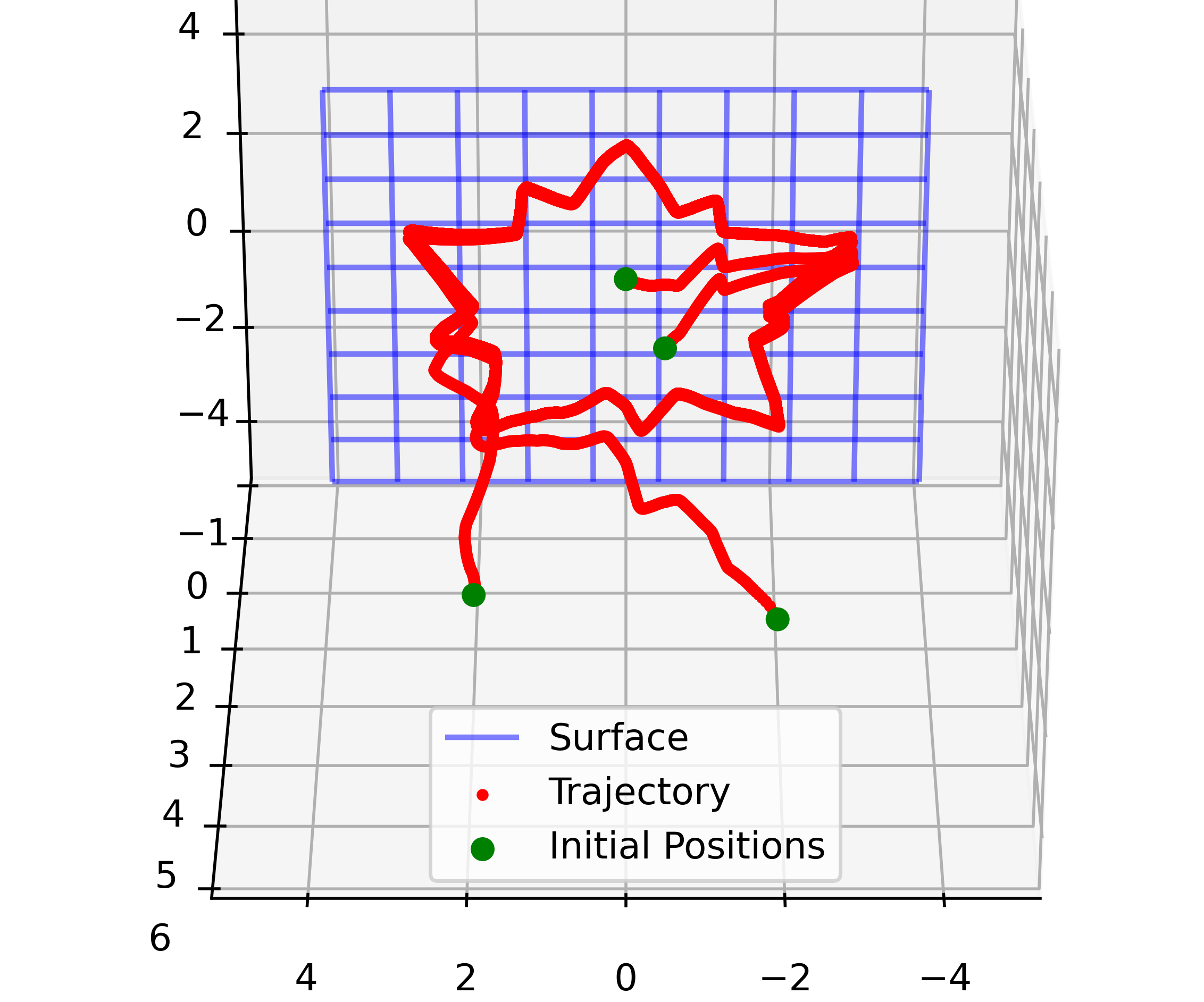}}  
\end{subfigure}%
\begin{subfigure}{.495\textwidth}
  \centering
  \fbox{
  \includegraphics[width=.33\linewidth]{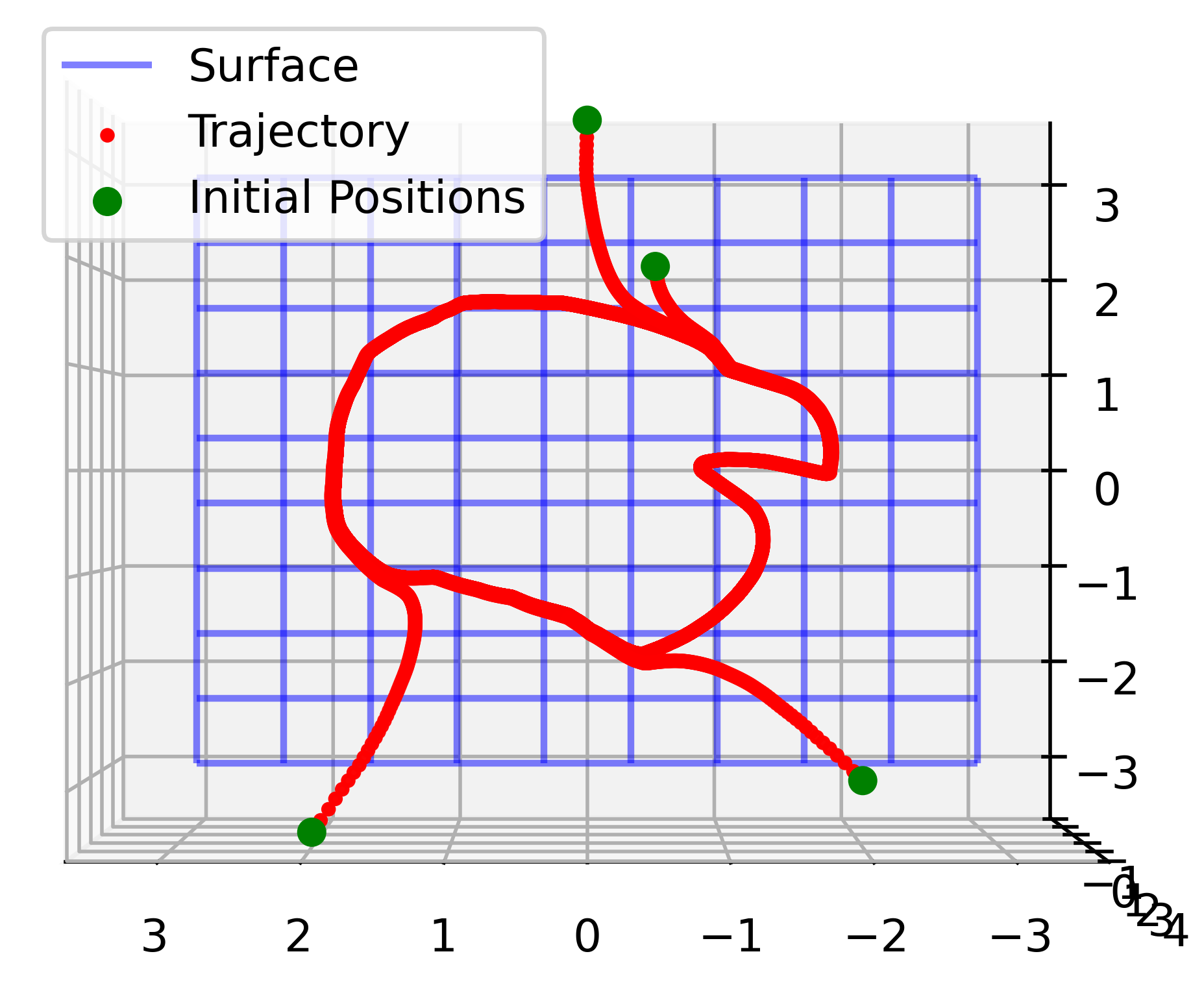}%
  \includegraphics[width=.33\linewidth]{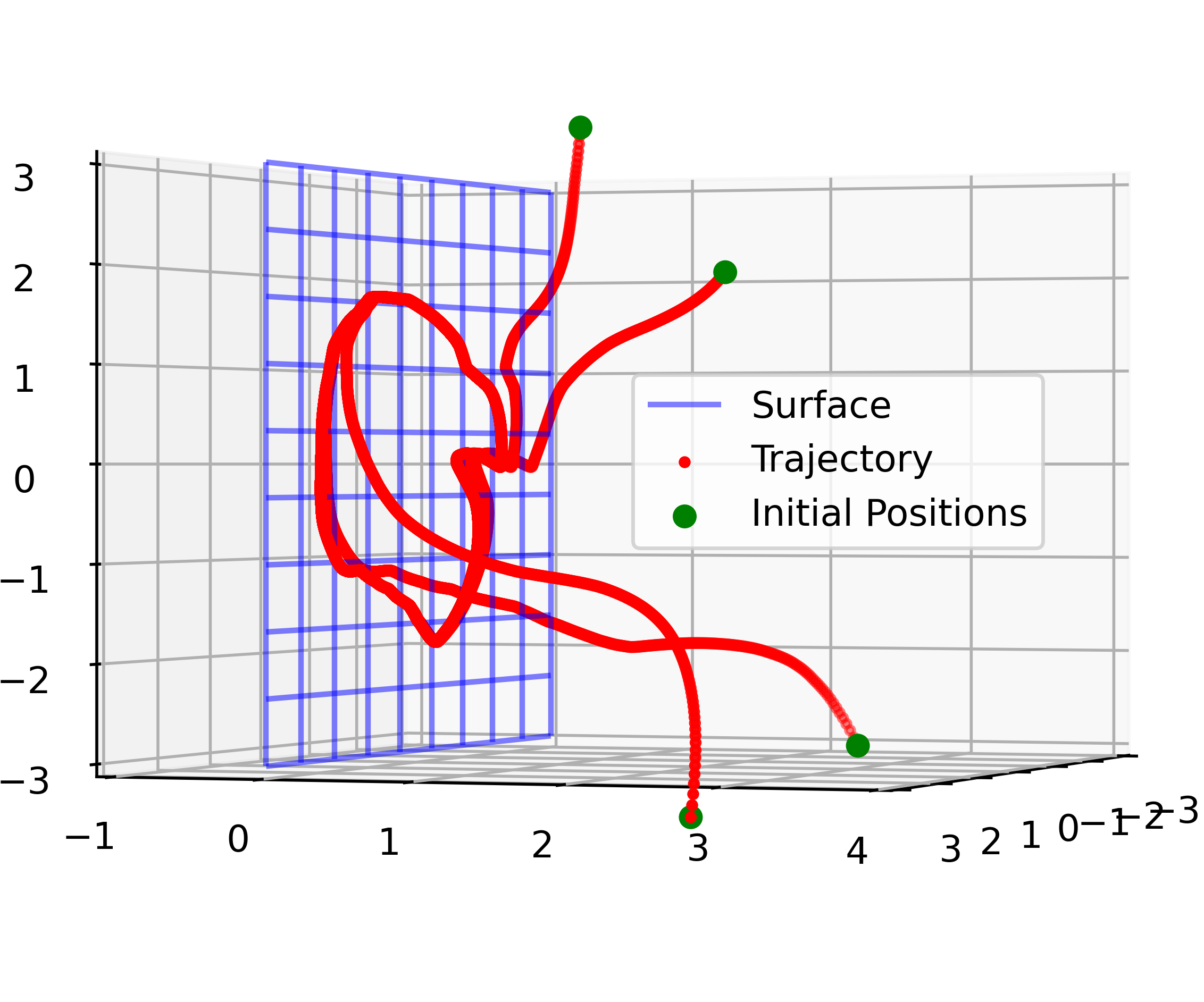}%
  \includegraphics[width=.33\linewidth]{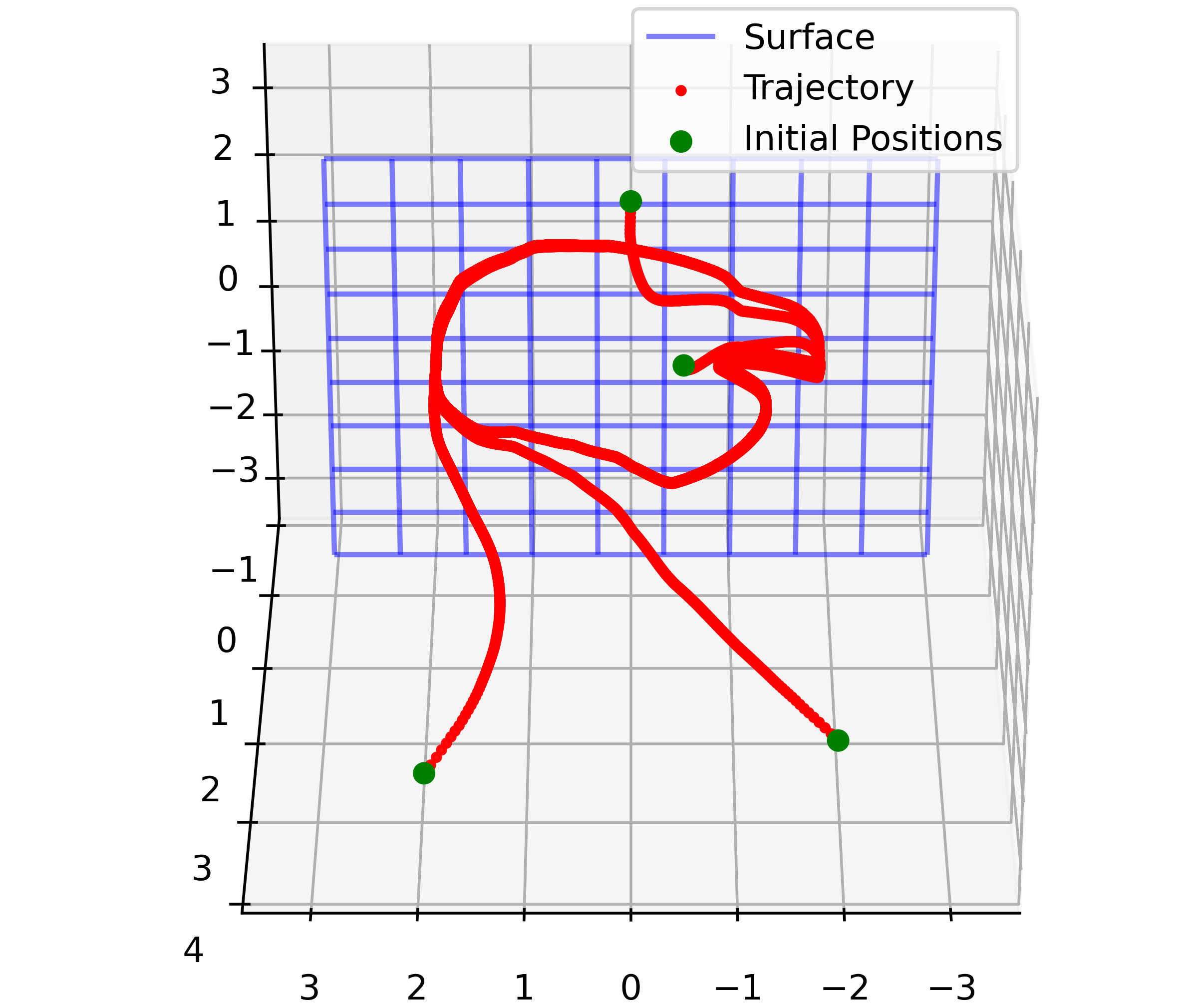}}  
\end{subfigure}%
\caption{Qualitative results of learning diffeomorphisms to shape the circular limit cycle into outlines of the \textbf{whale}, \textbf{dog}, \textbf{flower}, and \textbf{eagle}. We also show, at different viewing angles, of example trajectories integrated from multiple initial 3D positions (in green). We observe that each trajectory is able to converge onto the shaped limit cycle on the surface.}
\label{fig:shapes}
\end{figure*}

\subsection{Ray-tracing onto Surface}
After prompting the user to sketch the desired limit cycle on the surface on the camera image, we are assumed to have a set of $n$ 2D coordinates, $\bb{p}$, and corresponding depths, $d$, to the surface, i.e. $\{\bb{p}_{i},d_{i}\}_{i=1}^{n}$. We call the set of 2D coordinates the \emph{view-space shape}. We follow the~\cite{zhi2023learning} and assume a pin-hole camera. We construct a ray in 3D which passes through each 2D coordinate, $\bb{s}_{i}=\bb{o}+r(\bb{p}_{i})d_{i}$, where $\bb{o}$ and $r$ are camera origin and projection direction respectively. These are obtainable from the camera parameters and camera position. We collect the set of projected points, $\mathcal{S}=\{\bb{s}_{i}\}_{i=1}^{n}$, and use this as our training data. Note that as we align the flat surface to be the $x,z$-plane at $y=0$, by convention, we drop the $y$-axis of our projected points and simply consider the 2D coordinates of the sketch on the surface. \Cref{fig:example_ray_trace} shows an example of projecting a pentagon shape from view-space to a surface, with the traced rays in blue and the pentagon on the surface in red.

\subsection{Hausdorff Distance Loss}
The main component of the loss function is a measure of similarity between the shape specified by the user and the limit cycle. This requires us to define a distance between the set of sketched points projected onto the surface, $S$, and the limit cycle, $L$. Here, we compute a discretised Hausdorff distance~\cite{hausdorff1914grundzuge}, which provides distance between two point sets. Intuitively, the Haussdorff distance takes the larger of the maximum distances from one set to the other, and its reverse. A visualisation of this intuition is given in \cref{fig:example_HD}. This is defined as:
\begin{align}
H(S, L) =\!\max\!\left\{ \max_{\bb{s}\!\in S}\!\min_{\bb{l} \in L}\!D(\bb{s},\!\bb{l}),\!\max_{\bb{l}\!\in L} \!\min_{\bb{s} \in L}\!D(\bb{l},\!\bb{s})\!\right\},
\end{align}
where $D$ is a metric distance between individual elements, here, we simply use the $L2$ distance. 

To prevent the INN from learning diffeomorphisms that excessively distort the base system, we can regularise the diffeomorphism towards the identity function. This can be done by adding an additional regularisation term minimising the difference between $\bb{x}$ and $F^{-1}(\bb{x})$. This term can be computed via uniformly drawing $m$ samples, $\bb{\hat{x}_{1}},\ldots,\bb{\hat{x}_{m}}$, within a region of interest in state-space and evaluating their mean distances between $F^{-1}(\bb{\hat{x}_{1}}),\ldots,F^{-1}(\bb{\hat{x}_{m}})$. We arrive at the combined loss function:
\begin{align}
\ell=H(S, L)+\alpha \frac{1}{m}\sum_{i=1}^{m}\lvert\lvert\hat{x}_{i}-F^{-1}(\bb{\hat{x}_{i}})\lvert\lvert_{2}^{2},
\end{align}
where $\alpha$ controls the regularisation strength. We can then apply gradient-based optimisers, such as ADAM~\cite{Kingma2015AdamAM}, to learn the weights of the INN.

\subsection{Theoretical Guarantees on Learning Cycles}\label{sec:guarentees}
A particular question of interest is: What classes of 2D shapes can we find a diffeomorphism, to ``morph'' our limit cycle of the base system in \cref{eqn:base_sys}, at $y=0$ into? 

We show that any 2D shape that can be represented by a smooth and non-intersecting closed curve is diffeomorphic with the unit circle, and hence can, in theory, be ``morphed'' into by the limit cycle of our base system. 

\begin{prop}
Let $\mathcal{C}$ be a smooth and non-intersecting closed curve in $\mathbb{R}^{2}$. Then, $\mathcal{C}$ is diffeomorphic to the unit circle $\mathcal{S}^{1}:=\{(u,v)\in\mathbb{R}^{2}|u^2+v^2=1\}$.
\end{prop}\label{prop:diff_circle}
\begin{proof}
\textbf{Parameterise the curve, and extend to a periodic function}: Since $\mathcal{C}$ is a smooth curve in $\mathbb{R}^{2}$, we can find a parameterisation $\gamma:[0,1)\rightarrow\mathbb{R}^{2}$, such that $\gamma'(t)\neq0$ for $t\in[0,1)$. As $\mathcal{C}$ is closed, we can extend $\gamma$ to be periodic function on all of $\mathbb{R}$, such that $\gamma(t+1)=\gamma(t)$ for $t\in \mathbb{R}$. Additionally, as $\mathcal{C}$ is also non-intersecting, each point on $\mathcal{C}$ can be uniquely mapped to $t\in[0,1)$, hence $\gamma$ is a diffeomorphism. 

\textbf{Map $[0,1)$ to $\mathcal{S}^{1}$}: Furthermore, consider the mapping $\psi:\mathbb{R}\rightarrow\mathcal{S}^{1}$, given by $\psi(t)=\exp(2\pi it)$. This is a smooth map that wraps the real line around the circle infinitely many times and has a period of $1$. There exists a smooth inverse $\psi^{-1}$ which maps to $[0,1)$.

\textbf{Create the diffeomorphism between $\mathcal{C}$ and $\mathcal{S}^{1}$}: Consider the composition $\phi=\gamma^{-1}\circ\psi:\mathcal{C}\rightarrow [0,1) \rightarrow\mathcal{S}^{1}$. This is a diffeomorphism because it is a composition of two diffeomorphisms.
\end{proof}

Note that this proposition extends to a circle of arbitrary radius as the circle itself would be diffeomorphic to the unit circle. Moreover, the authors of~\cite{coupling_universal} show that coupling-based INNs are \emph{universal diffeomorphism approximators}. Informally, this means that they can approximate any diffeomorphism, $\phi$, to arbitrary accuracy.

\begin{figure}[t]
    \centering
    \fbox{\includegraphics[width=0.12\textwidth]{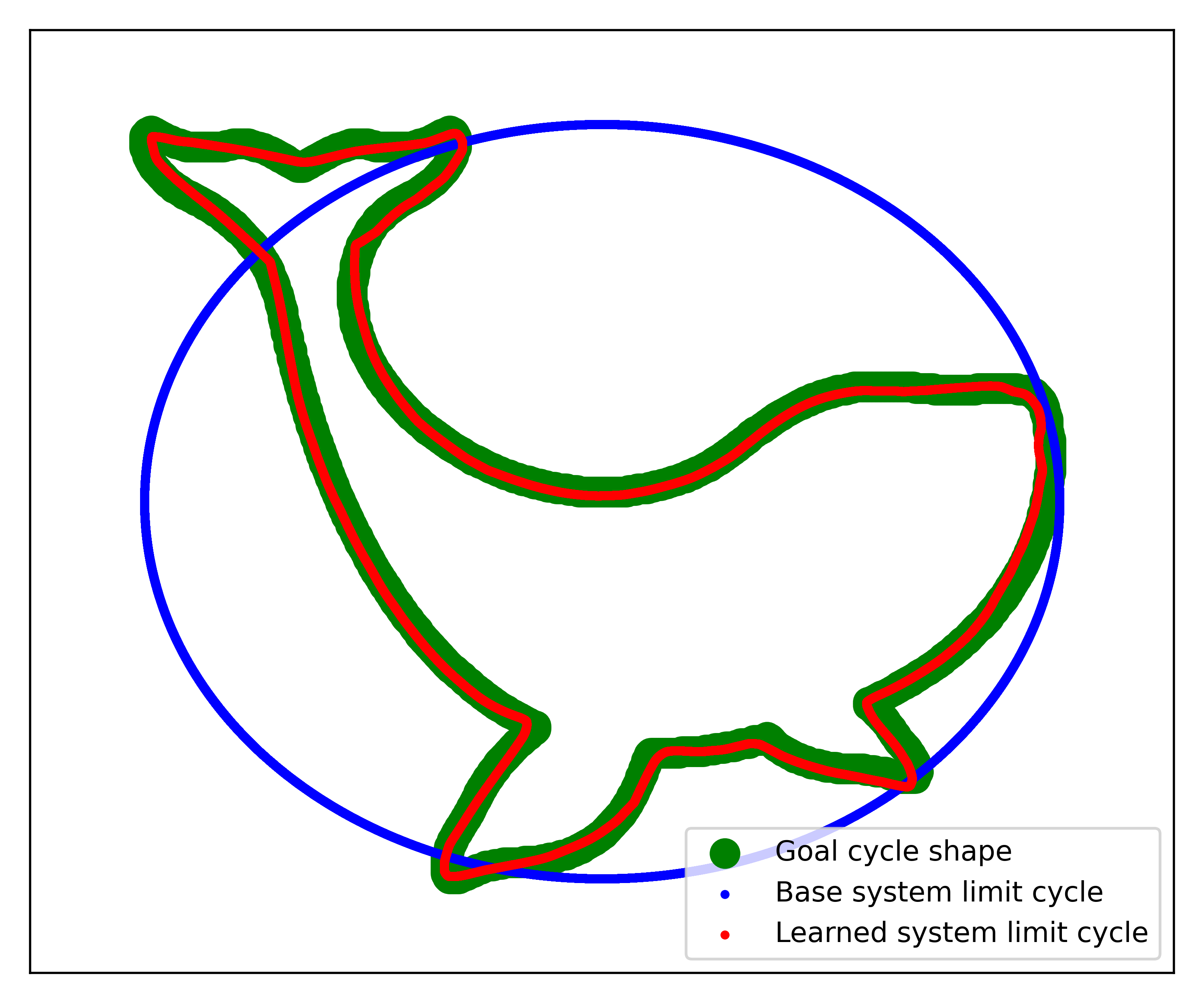}}%
    \fbox{\includegraphics[width=0.12\textwidth]{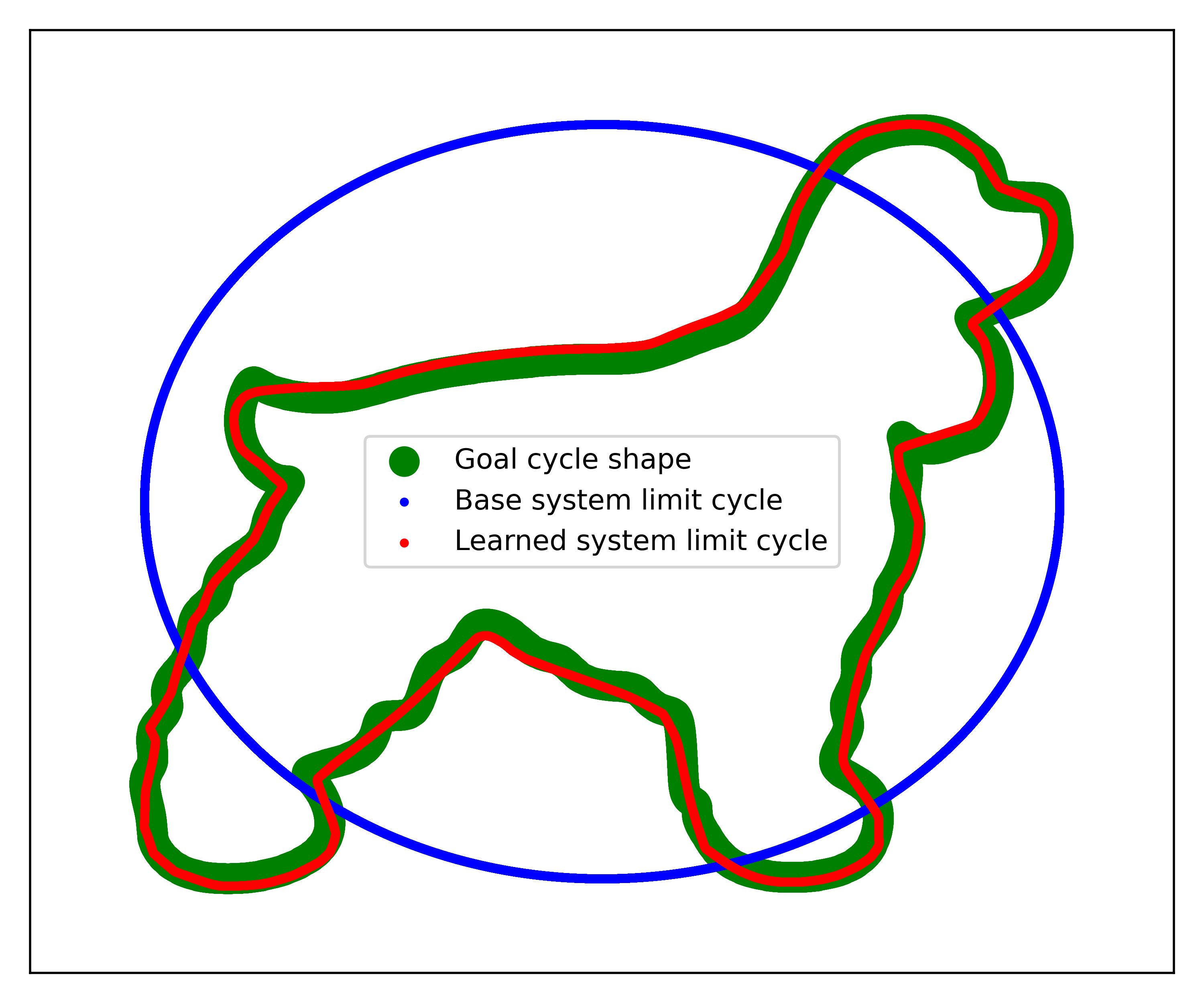}}%
    \fbox{\includegraphics[width=0.12\textwidth]{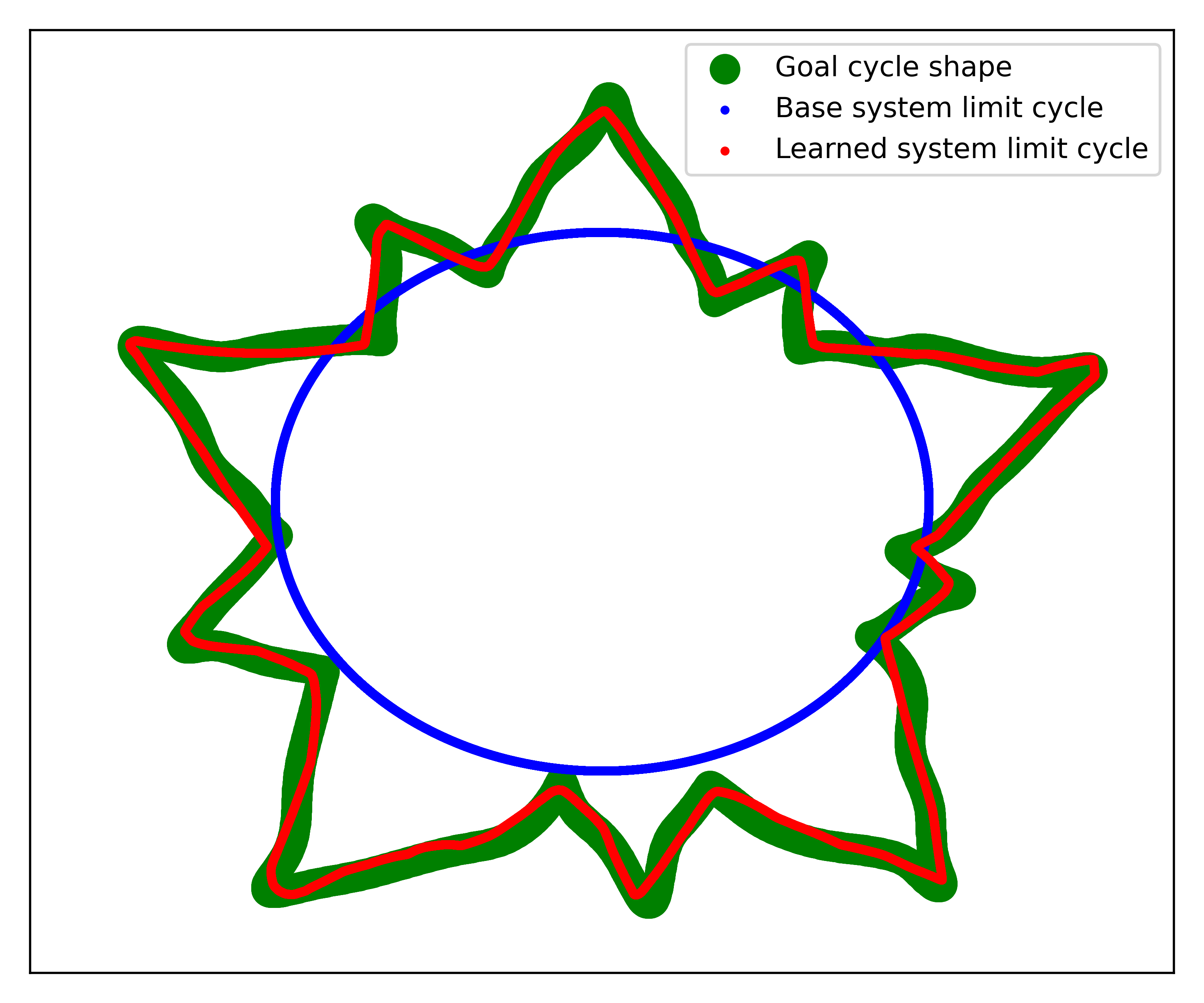}}%
    \fbox{\includegraphics[width=0.12\textwidth]{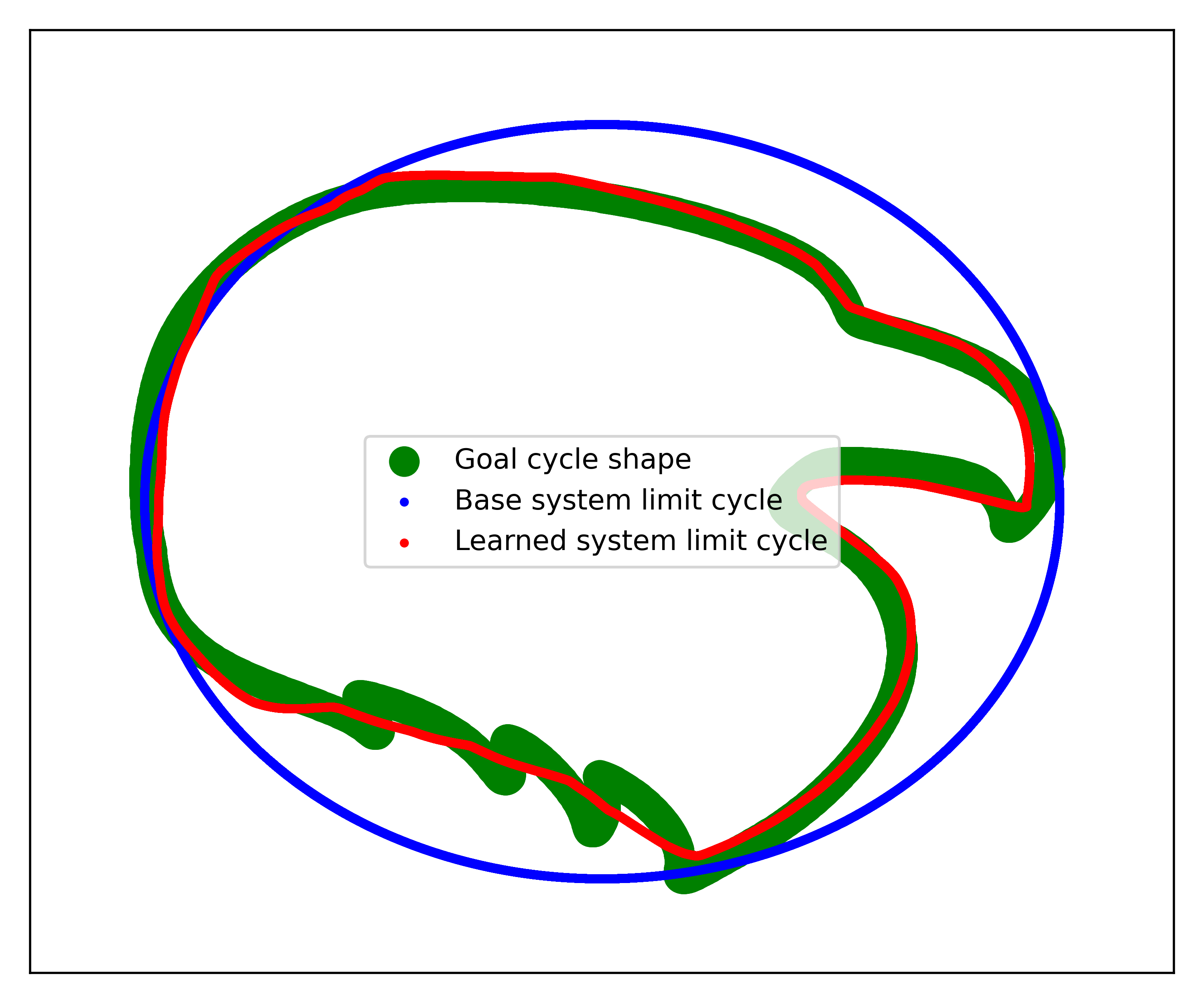}}%
    \caption{Our learned cycles on the $x,z$-plane. Cycle of the base in blue, the target data in green, and the cycle learned in red.}\label{fig:2d_result}
\end{figure}

\begin{figure}[t]
\setlength{\fboxsep}{0pt}
\centering
  \centering
\fbox{\includegraphics[width=.32\linewidth]{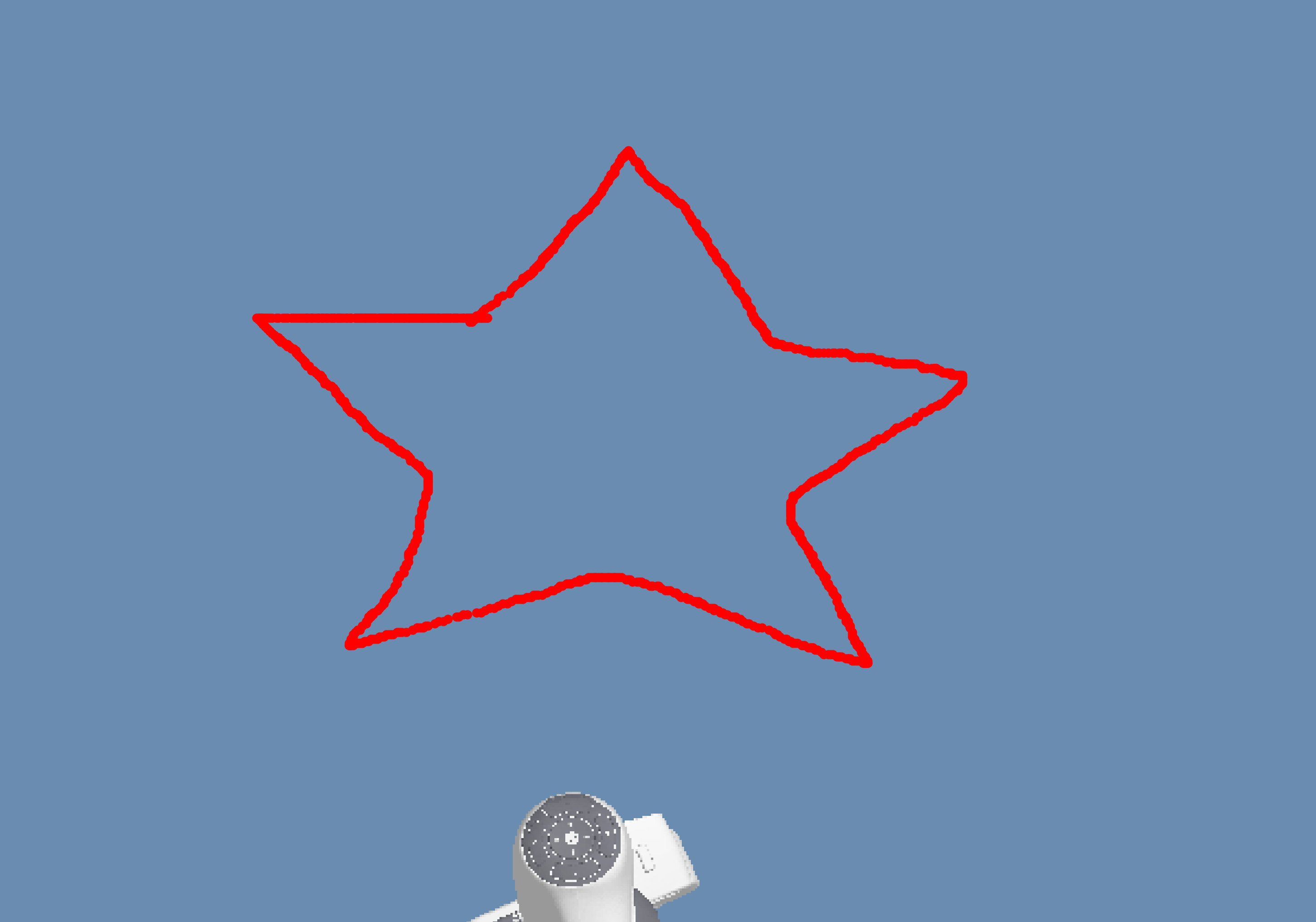}}%
\fbox{\includegraphics[width=.32\linewidth]{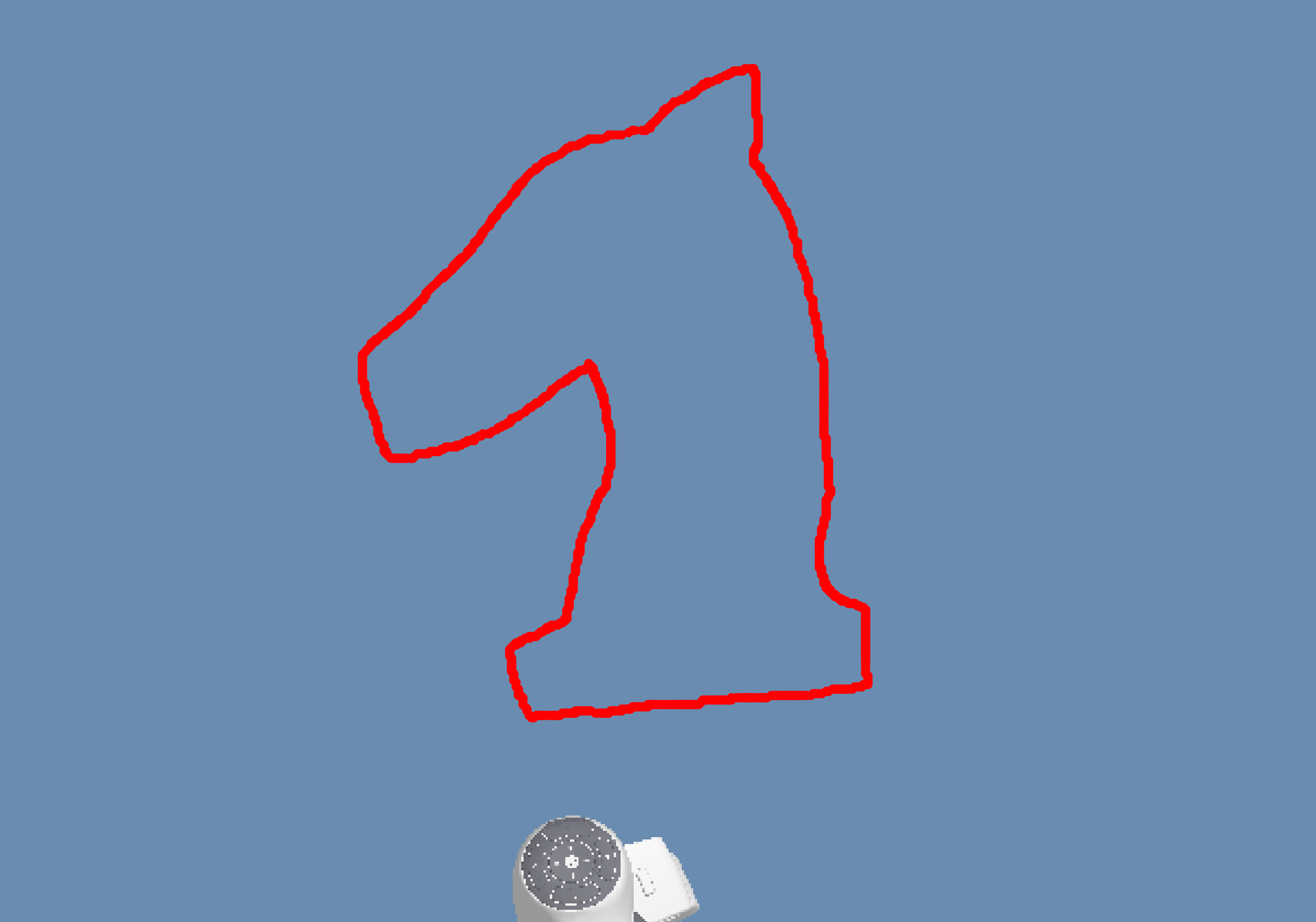}}%
\fbox{\includegraphics[width=.32\linewidth]{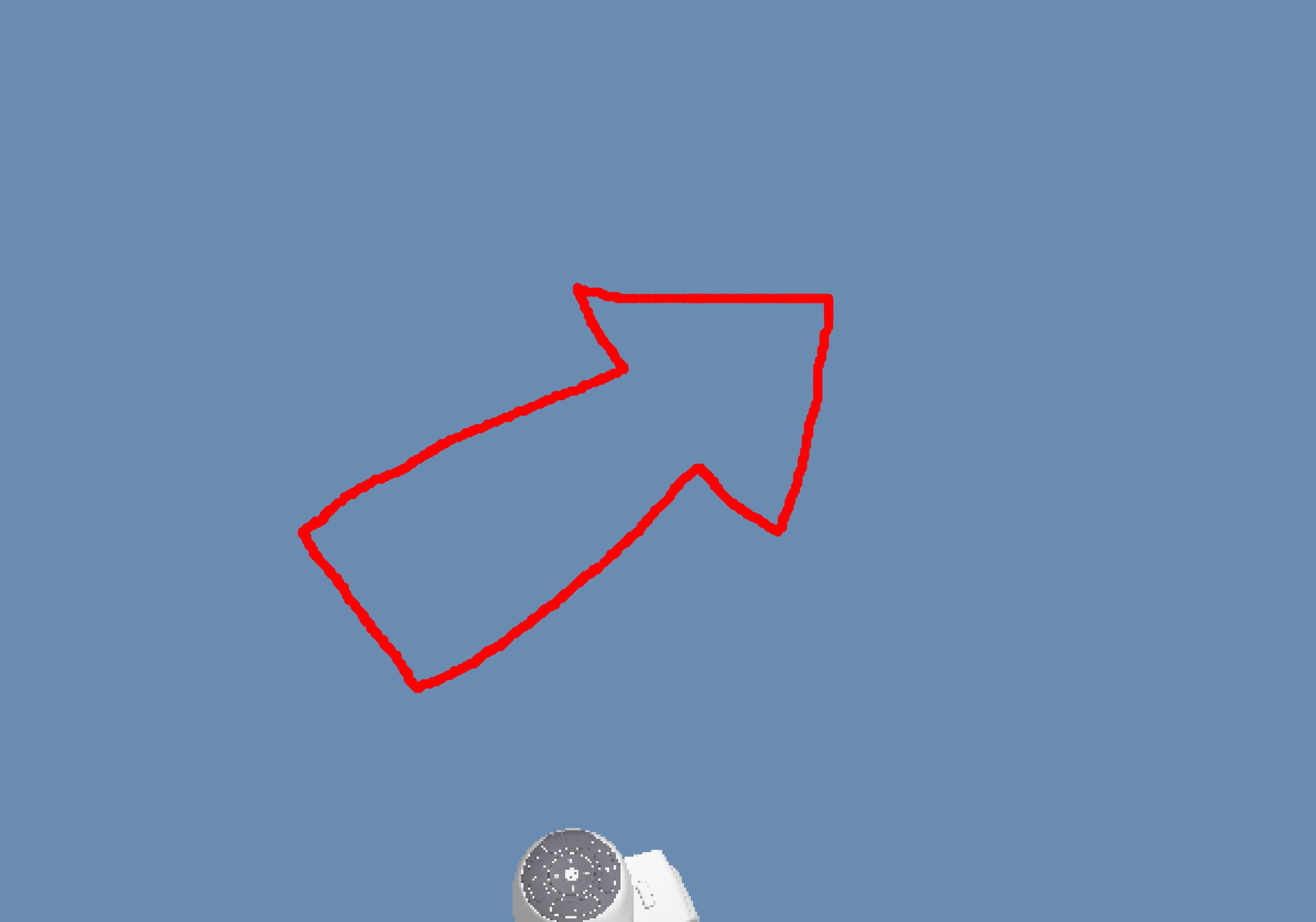}}  

\fbox{\includegraphics[width=.32\linewidth]{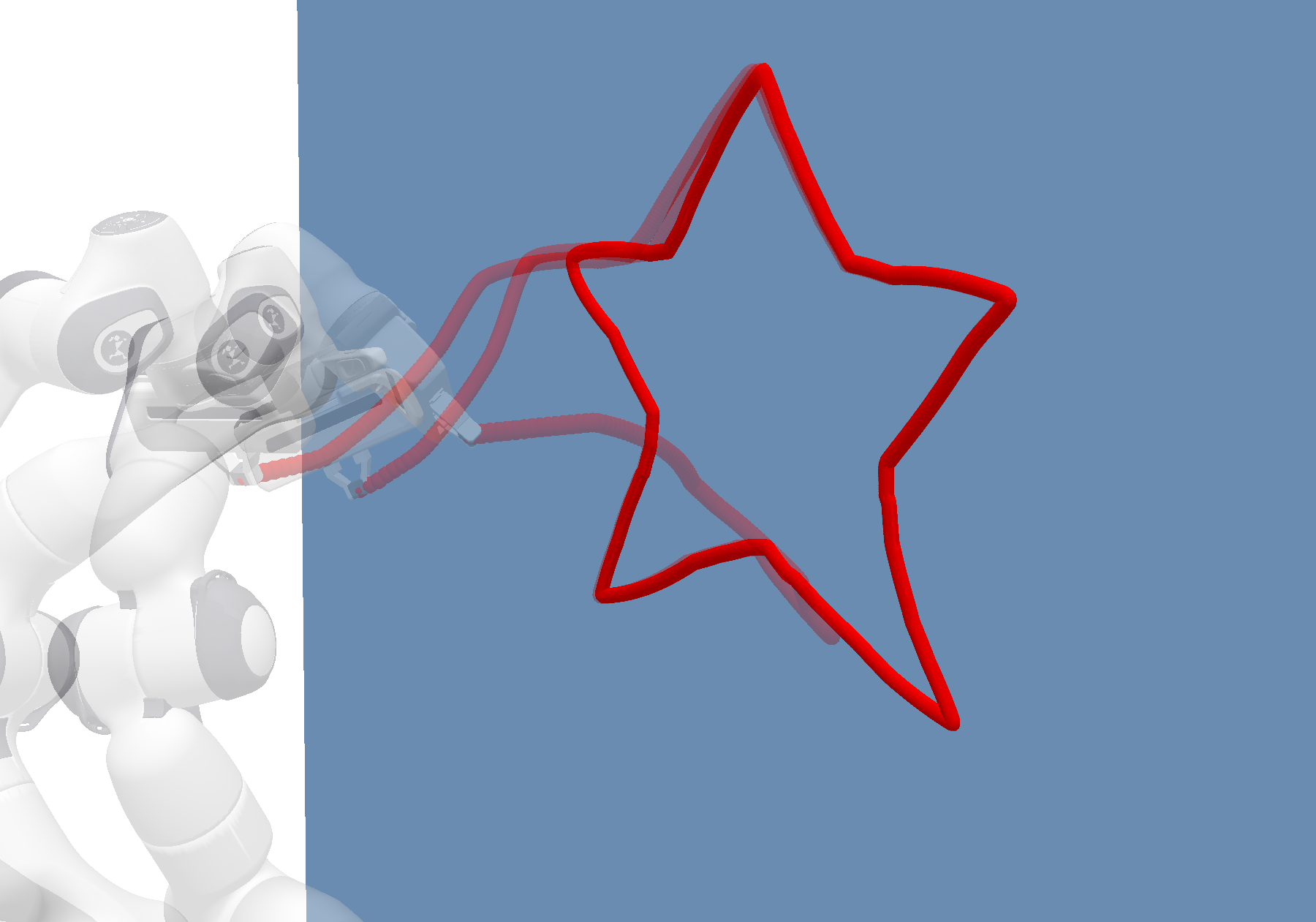}}%
\fbox{\includegraphics[width=.32\linewidth]{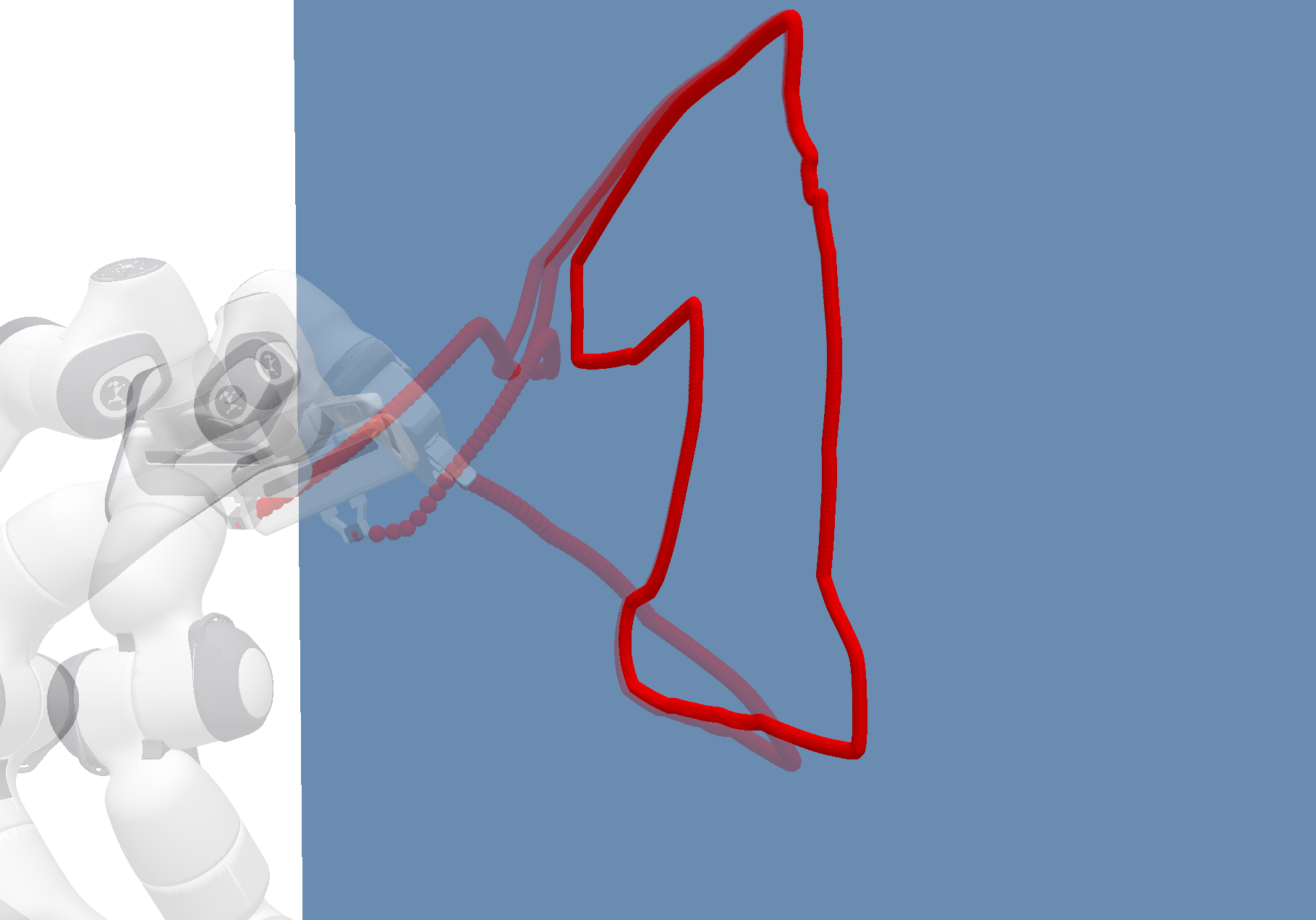}}%
\fbox{\includegraphics[width=.32\linewidth]{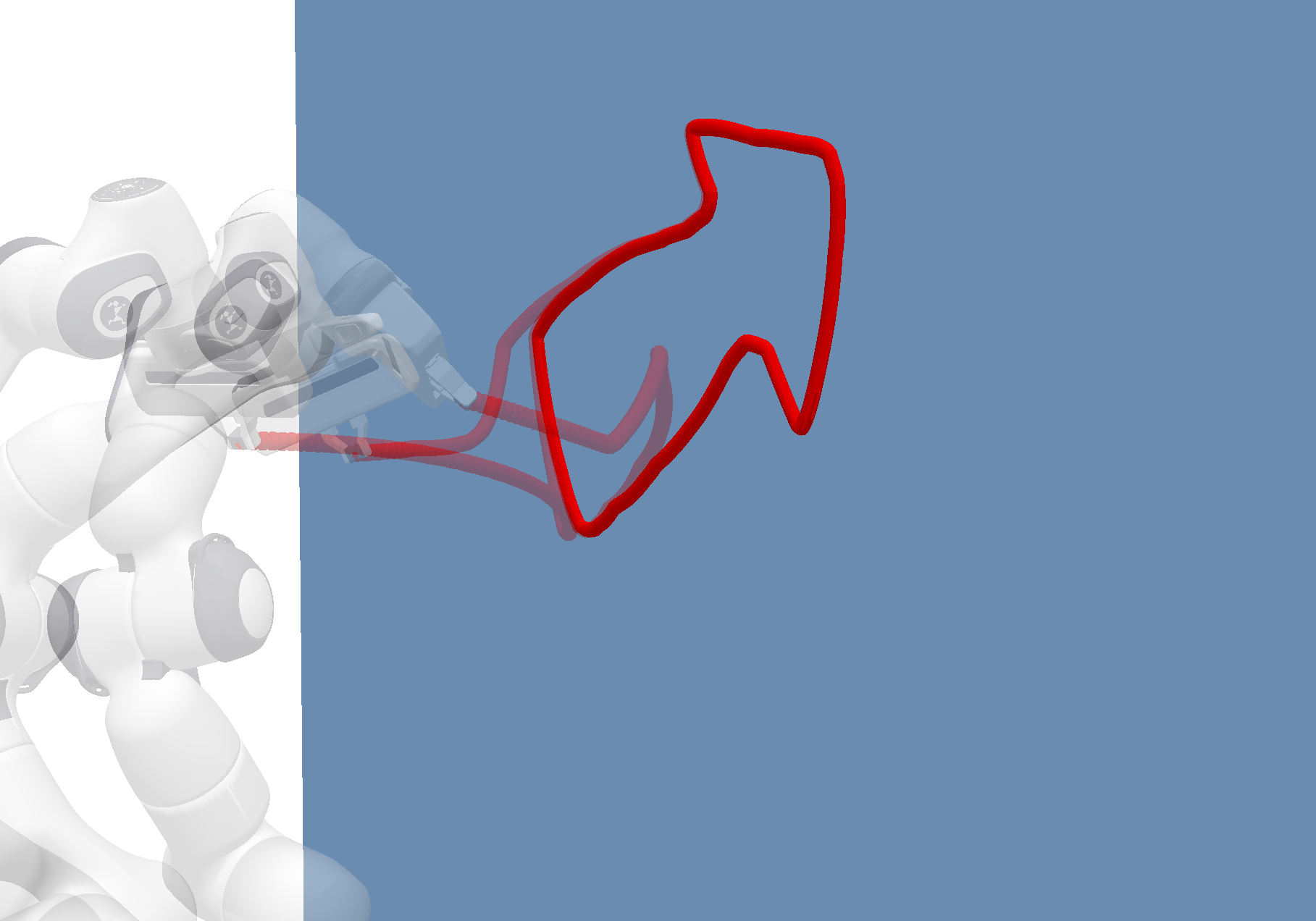}}  
\caption{We evaluate SDDT in the PyBullet Simulator. (top) The user is provided a view from a camera in the scene and is prompted to sketch a star, a knight chess piece, and an arrow on the image respective. The user sketches (in red) are overlaid over the camera image. (bottom) For each different sketch, we generate motion trajectories from SDDT from 3 different robot configurations. We observe that each of these trajectories is able to consistently and accurately converge onto the desired shape.}
\label{fig:Results_SDDT}
\end{figure}

\begin{figure}[h]
\centering
\fbox{\includegraphics[width=0.42\textwidth]{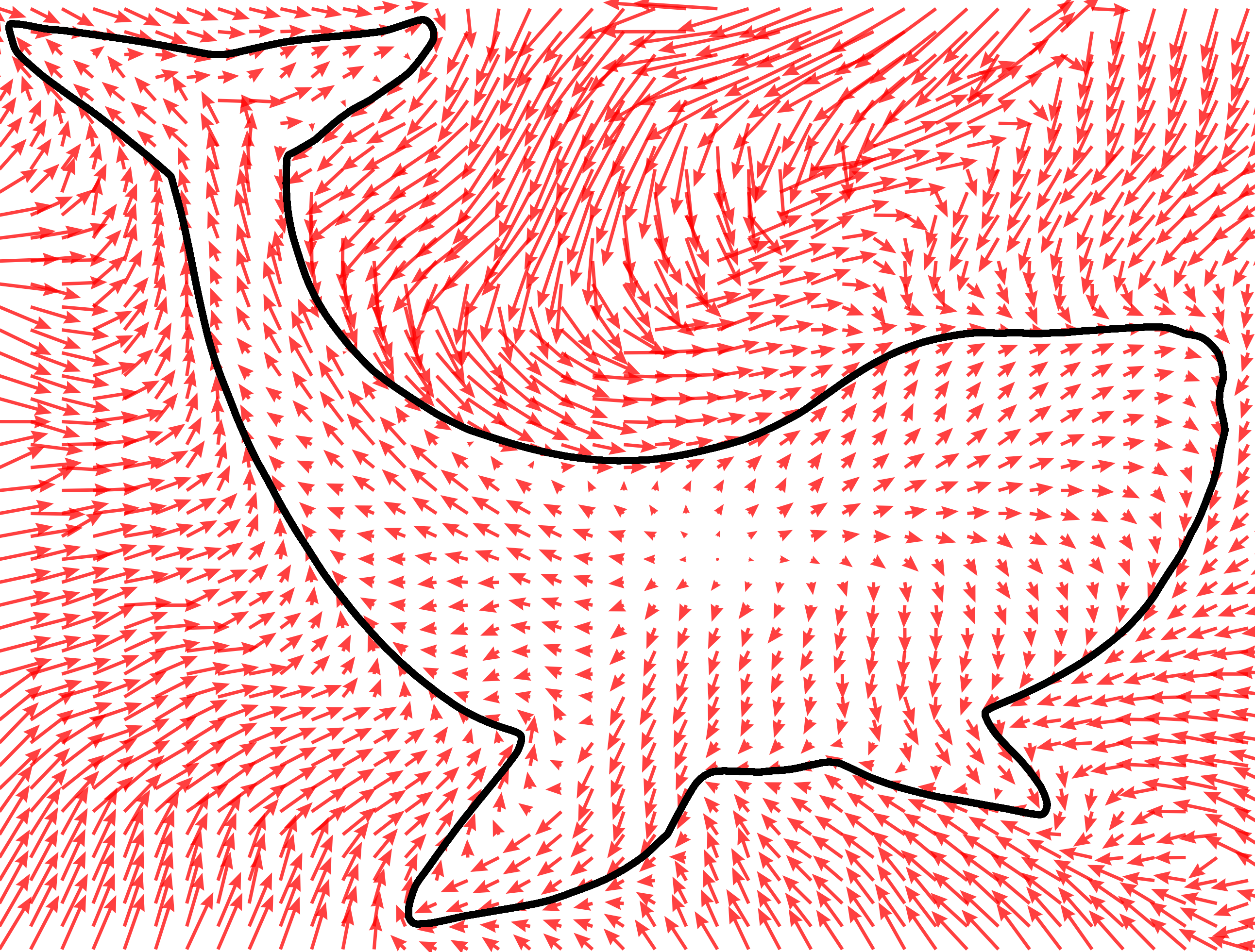}}
\caption{The learnt O.A.S. vector field on the $x,z$-plane. The vector field (red arrows) pushes points off the cycle onto the stable cycle (in black).}
\label{fig:vector_field}
\end{figure}

\section{Experimental Results}\label{sec:experiments}
We experimentally evaluate the performance and test the robustness of our proposed SDDT method, both in simulation and on real-world robots. We begin by empirically investigating the learning capacity of SDDT, and stress-testing it to learn complicated limit cycles (\cref{sec:intricate}). We then highlight the benefits of enforcing O.A.S. by comparing SDDT against neural ODEs~\cite{NODEs}, which parameterise the dynamics of the system as a free-form neural network (\cref{sec:intricate}). Lastly, we demonstrate the applicability of SDDT in the real world by deploying the framework on a quadruped-mounted manipulator. 

\subsection{A Qualitative Analysis: Learning Challenging Cycles}\label{sec:intricate}
We seek to explore the capabilities of our proposed method for learning systems with limit cycles that are intricate and vary greatly from the circular limit cycle of the base dynamical system. Here, we extract silhouette outlines and investigate how well SDDT is able to ``morph'' the cycle into the outlines. 

\begin{table}[t]
    \centering
    \caption{Performance of SDDT and baselines, on each task, as measured by Hausdorff distance (lower is better).}
    \label{tab:comparison}
    \setlength{\heavyrulewidth}{0.2em}
    \begin{tabular}{l|cccc}

        \toprule
        \hline
            & O.A.S. & Star & Knight & Arrow \\
        \midrule
        SDDT (ours)  & \cmark & \textbf{0.011} & \textbf{0.011} & \textbf{0.017} \\
        Neural ODEs~\cite{NODEs}  & \xmark & 0.040  & 0.035 & 0.063 \\
        Base System  & \cmark & 0.133 & 0.201 & 0.209 \\
        \hline
    \bottomrule
    \end{tabular}
\end{table}

\begin{figure*}[t]
    \centering
    \fbox{\includegraphics[width=0.242\textwidth]{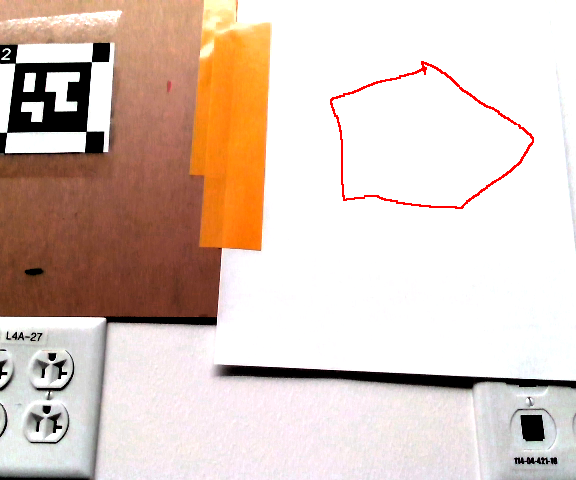}}%
    \fbox{\includegraphics[width=0.242\textwidth]{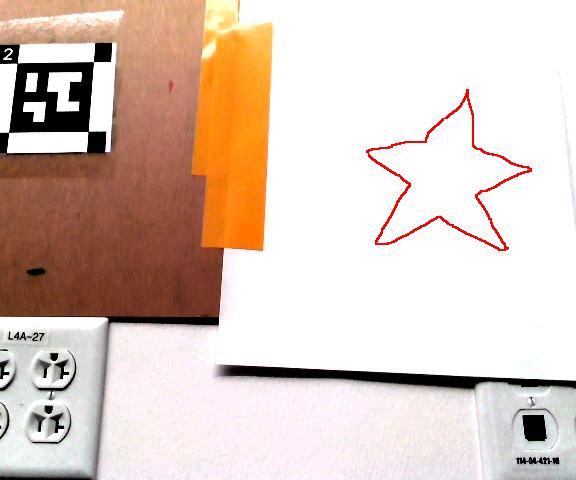}}%
    \hspace{0.05em}
    \fbox{\includegraphics[width=0.242\textwidth]{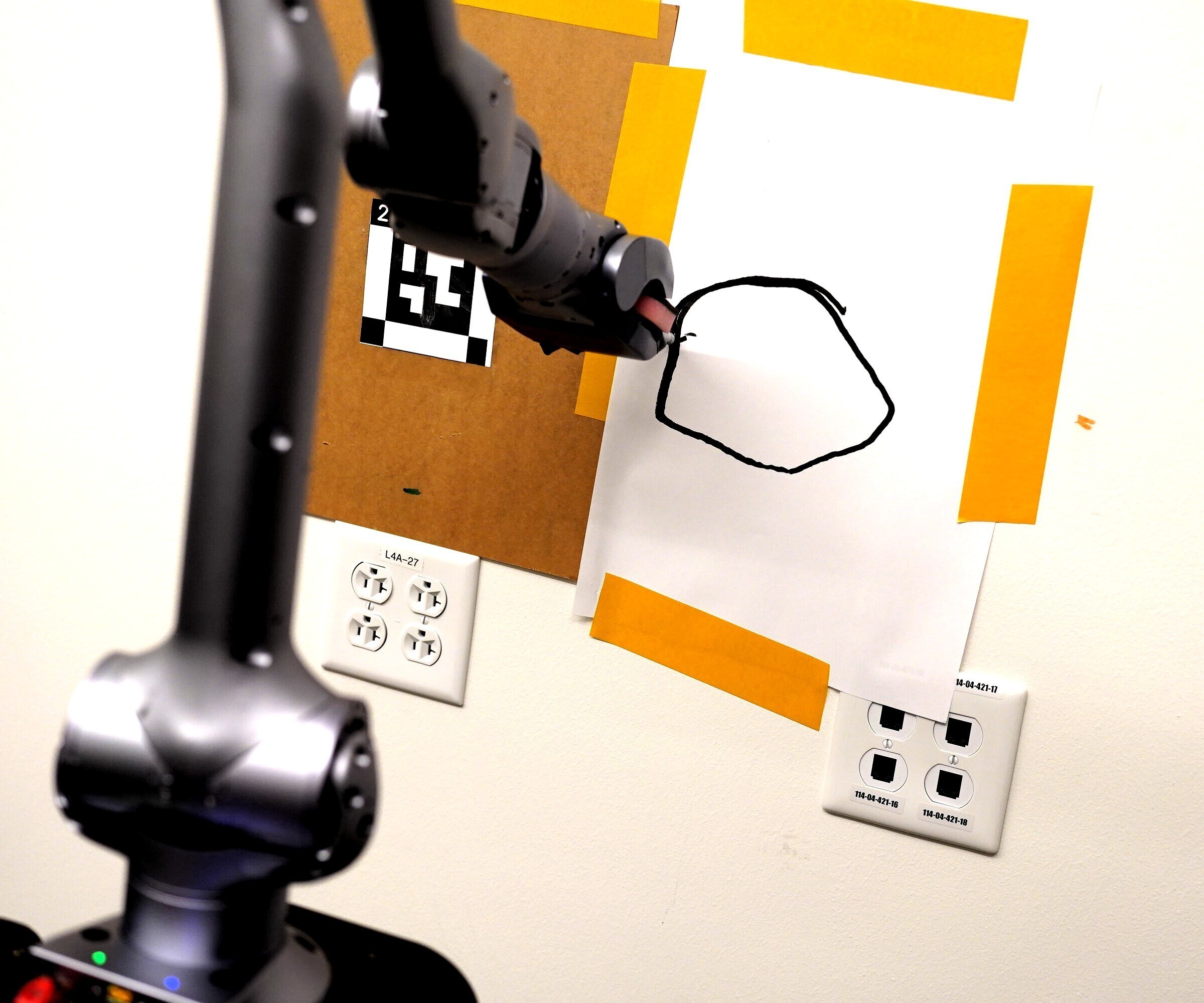}}%
    \fbox{\includegraphics[width=0.242\textwidth]{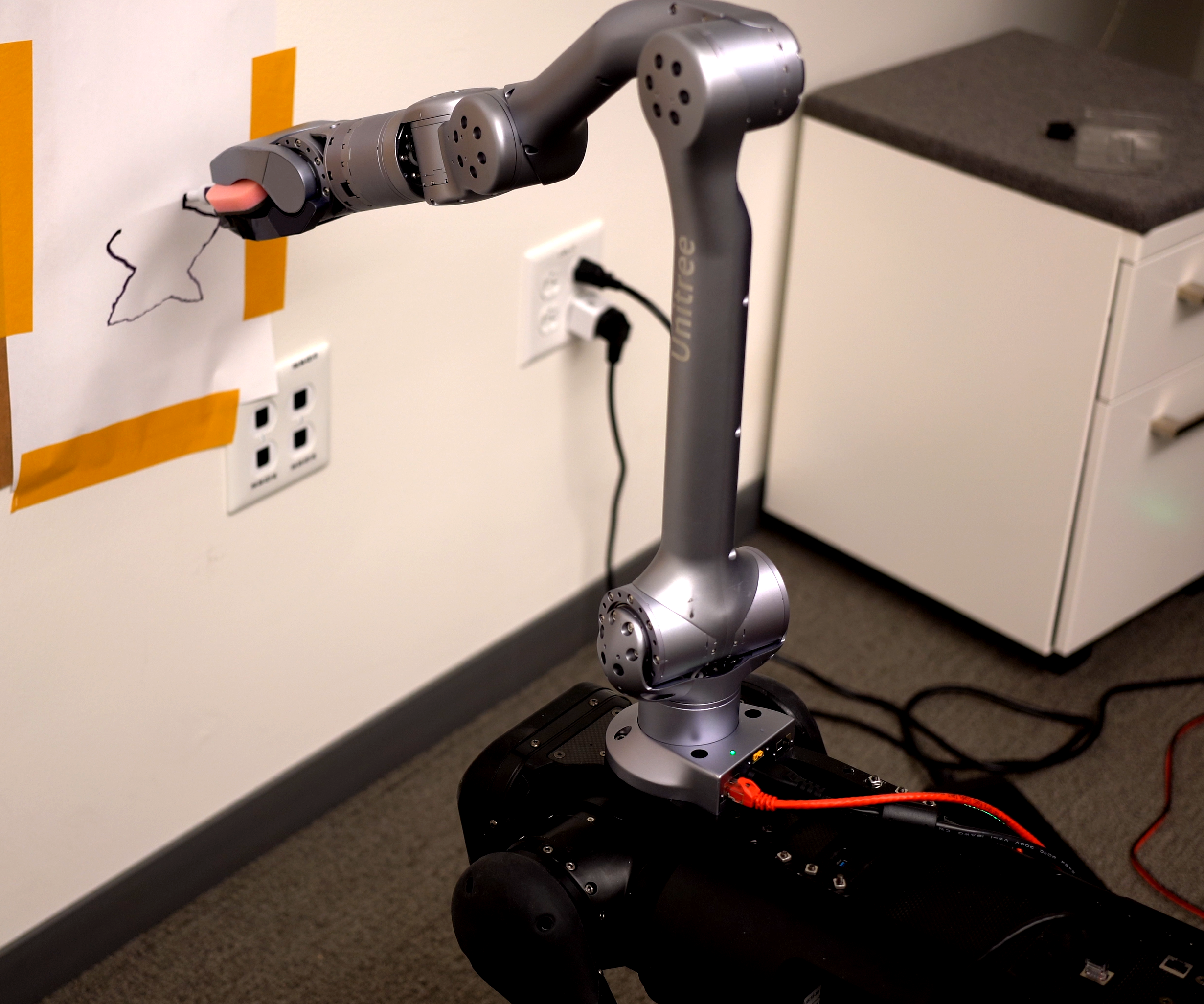}}%
    
    \caption{SDDT is particularly useful when egocentric images, from onboard cameras, are available. We run our real-world experiments on a quadruped with a mounted arm. (Left) We sketch the shapes of a pentagon and a star (in red) on an egocentric view from the onboard camera. (Right) The robot converges to the surface, stabilises at, and traces out the diagrammatically provided shapes.}\label{fig:real_robots}
\end{figure*}

We use outlines of a \textbf{whale}, a \textbf{dog}, a \textbf{flower}, and an \textbf{eagle} and learn an INN to morph the base system's limit cycle to match the outline. Throughout this paper, we use the INN models in the \emph{FrEIA} library~\cite{Freia}, which is built on \emph{PyTorch}~\cite{pytorch}. In \cref{fig:shapes}, we provide qualitative results of both the limit cycle and trajectories integrated from the resulting dynamical system. We observe that the limit cycle is generally able to be accurately shaped into each of these outlines, and the integrated 3D trajectories are able to approach the surface and smoothly converge onto the limit cycle. We also visualise and compare the original base system and the target 2D shape, on the $x,z$-plane in \cref{fig:2d_result}.

To gain insight into how the diffeomorphism ``morphs'' the $x,y$-plane at $y=0$, in \cref{fig:analysis}, we visualise additional quantitative results on the \textbf{whale} outline. On the left, we show a transport map showing examples of points on the orbit of the base system mapped to the learned system, with correspondence between points on the systems indicated by the grey line. On the right, we visualise the result of diffeomorphism operating on concentric circles (in red), starting from a radius of $0.1$ to $2$. This gives us an intuition of how the ambient space is stretched and compressed to match the target shape (in black). In \cref{fig:vector_field}, we show the vector field of the resulting learned system, with the limit cycle shown in black. We observe that points not in the vector field shall be attracted towards the stable cycle --- points inside the cycle push outwards while those outside push inwards.   

\subsection{SDDT outperforms Both Free-form Neural ODEs and the Restrictive Base System}
We evaluate, in the PyBullet Simulator~\cite{coumans2019}, the performance of the SDDT framework. We simulate a Franka on a mobile base facing a wall, with an RGB-D camera positioned behind the robot. We seek to generate robot motion that approaches the wall and converges onto the shape diagrammatically provided by the user on the camera image. We compare against the following baselines: 
\begin{enumerate}
    \item \textbf{Neural ODEs}~\cite{NODEs}: Neural ODEs learn dynamical systems by parameterising the dynamics as a neural network and then train on data. We use a Neural ODE to learn the dynamics on the $x,z$-plane, and retain the attractor towards the surface in the $y$-axis direction. The dynamics of Neural ODEs are completely free-form, with no assumption made on stability.
    
    \item \textbf{Base System} (defined in \cref{eqn:base_sys}): We evaluate how well our stable base system performs. We allow the radius and the origin of the base system to be tuned, such that the limit cycle minimises the distance between the data. The base system is highly structured and contrasts with the entirely learning-based Neural ODE.
\end{enumerate}

After collecting sketches of a \textbf{star}, a \textbf{knight}, and an \textbf{arrow}, we train each of these models and integrate trajectories at three different initial robot configurations. Here, we use the implementation of Neural ODEs provided in the \emph{torchdiffeq} library~\cite{NODEs}. To measure how well the trajectories from the traced motion match the diagrammatic sketch provided by the user, we take points that have $y$-values under $10^{-4}$ to be in contact with the surface, and compute the Hausdorff distance between the trajectory and sketch ray-traced onto the surface. A qualitative evaluation of our generated trajectories, as well as the user-provided sketches, can be seen in \cref{fig:shapes}, with results provided in \cref{tab:comparison}. We observe that SDDT imbues knowledge of stability into the system via enforcing O.A.S., and outperforms Neural ODEs which treat the dynamics as a black-box. SDDT is also sufficiently flexible to learn complex patterns, greatly outperforming the inflexible stable base system. 

\subsection{SDDT on Real Robots}
We demonstrate the applicability of SDDT on real-world robots, by applying SDDT on a \emph{Unitree Aliengo} quadruped with an attached 6-DOF \emph{Z}1 manipulator. The user is shown egocentric views of the environment via the RGB-D camera on board the quadruped and is asked to sketch a \textbf{pentagon} and a \textbf{star}. SDDT is then used to learn stable systems shaped by the projection of the drawing. We then integrate a trajectory from the current end-effector position and track the trajectory with a marker pen to trace out the corresponding motion patterns. We observe that in both instances the quadruped mount manipulator was able to approach the surface and stabilise on a cycle that matched the diagrammatically specified shapes, despite minor inaccuracies introduced by contact forces. In \cref{fig:real_robots}, we overlay the provided sketches onto the egocentric view images from the quadruped and show the manipulator converging onto the surface and tracing the desired shapes. 

\section{Conclusions and Future Work}\label{sec:conclusion}
In this work, we tackle the problem of learning robot policies that approach a surface and trace on it periodically. Robot motions of this kind are applicable to painting, wiping, and sanding tasks. We take a \emph{diagrammatic learning} approach, where the type of periodical pattern is provided by the user providing a 2D sketch on an image of the scene. We contribute the novel \emph{Stable Diffeomorphic Diagrammatic Teaching} (SDDT) method, where ray-tracing is used to project the user's sketch onto the surface, and an Orbitally Asymptotically Stable (O.A.S.) dynamical system, which converges to a cyclic orbit, is learned as a policy for the robot's motion. SDDT learns O.A.S. systems by learning a diffeomorphism that morphs a known stable base system into the desired system. We provide theoretical insight into the classes of 2D shapes the stable limit cycle can be shaped into, and provide extensive empirical evaluations of SDDT, both in simulation and on a real-world quadruped with a mounted manipulator. Future avenues of research include: (1) extending SDDT beyond flat surfaces, and to ensure stable motion patterns of curved surfaces; (2) extending SDDT to allow the specification of forces applied onto the surface.  

\bibliographystyle{ieeetr} 
\bibliography{bib}
\end{document}